\documentclass{article} 
\usepackage{PRIMEarxiv}

\usepackage[utf8]{inputenc} 
\usepackage[T1]{fontenc}    
\usepackage{hyperref}       
\usepackage{url}            
\usepackage{booktabs}       
\usepackage{amsfonts}       
\usepackage{nicefrac}       
\usepackage{microtype}      
\usepackage{lipsum}
\usepackage{graphicx}
\usepackage{tablefootnote}
\graphicspath{{media/}}     
\usepackage[utf8]{inputenc} 
\usepackage[T1]{fontenc}    
\usepackage{hyperref}       
\usepackage{url}            
\usepackage{booktabs}       
\usepackage{amsfonts}       
\usepackage{nicefrac}       
\usepackage{microtype}      
\usepackage{xcolor}         
\usepackage[acronym]{glossaries}
\usepackage{float}
\usepackage{xcolor}
\usepackage{comment}
\usepackage{amsthm} 
\usepackage{bm} 
\usepackage{bbm}
\usepackage{amssymb}
\usepackage{graphicx} 
\usepackage[acronym]{glossaries}
\usepackage[linesnumbered,ruled,vlined]{algorithm2e}
\usepackage{algpseudocode}
\usepackage{mathtools}
\usepackage{natbib}
\usepackage{footnote}
\usepackage{caption}
\makesavenoteenv{tabular}

\newtheorem{definition}{Definition}[section]
\newtheorem{proposition}[definition]{Proposition}

\newtheorem{lemma}[definition]{Lemma}
\newtheorem{observation}[definition]{Observation}
\newtheorem{assumption}[definition]{Assumption}
\newtheorem{remark}[definition]{Remark}

\makeglossaries
\newacronym{acr:per}{PER}{Prioritized Experience Replay}
\newacronym{acr:reaper}{ReaPER}{Reliability-adjusted Prioritized Experience Replay}
\newacronym{acr:rl}{RL}{Reinforcement Learning}
\newacronym{acr:dqn}{DQN}{Deep Q-Network}
\newacronym{acr:ddqn}{DDQN}{Double Deep Q-Network}
\newacronym{acr:tql}{TQL}{Tabular Q-learning}
\newacronym{acr:td}{TD}{Temporal Difference}
\newacronym{acr:tde}{TDE}{Temporal Difference Error}
\newacronym{acr:perg}{PER-g}{Greedy Prioritized Experience Replay}
\newacronym{acr:reaperg}{ReaPER-g}{Greedy Reliability-adjusted Prioritized Experience Replay}
\newacronym{acr:mdp}{MDP}{Markov decision process}

\title{Reliability-Adjusted Prioritized Experience Replay\footnote{}
}

\author{%
  Leonard S.~Pleiss\\
  Technical University Munich\\
  Munich, 80331 \\
  \texttt{leonard.pleiss@tum.de} \\
  \And
  Tobias Sutter \\
  University St. Gallen \\
  St. Gallen, 9000 \\
  \texttt{tobias.sutter@unisg.ch} \\
  \And
  Maximilian Schiffer \\
  Technical University Munich\\
  Munich, 80331 \\
  \texttt{schiffer@tum.de} \\
}

\begin{document}
\newcommand{\idxCurrent}{c}
\newcommand{\idxPrimary}{t}
\newcommand{\idxSecondary}{i}
\newcommand{\idxTertiary}{m}
\newcommand{\idxOracle}{l}
\newcommand{\idxSelected}{j}

\maketitle

\begin{abstract}
Experience replay enables data-efficient learning from past experiences in online reinforcement learning agents. Traditionally, experiences were sampled uniformly from a replay buffer, regardless of differences in experience-specific learning potential. In an effort to sample more efficiently, researchers introduced \gls{acr:per}. In this paper, we propose an extension to \gls{acr:per} by introducing a novel measure of temporal difference error reliability.
We theoretically show that the resulting transition selection algorithm, \gls{acr:reaper}, enables more efficient learning than \gls{acr:per}. We further present empirical results showing that \gls{acr:reaper} outperforms  both uniform experience replay and \gls{acr:per} across a diverse set of traditional environments including several classic control environments and the Atari-10 benchmark, which approximates the median score across the Atari-57 benchmark within one percent of variance.
\end{abstract}

\section{Introduction}
\gls{acr:rl} agents improve by learning from past interactions with their environment. A common strategy to stabilize learning and improve sample efficiency is to store these interactions -- called transitions -- in a replay buffer and reuse them through experience replay to increase sample-efficiency. When using experience replay, the agent obtains mini-batches for training by sampling transitions from the replay buffer. Mini-batches are traditionally obtained using random sampling. However, a prioritization schemes can help to select more informative transitions, improving convergence speed and, ultimately, agent performance significantly (\cite{schaul_prioritized_2015}). As such, the sampling scheme constitutes a performance-crititcal component for modern reinforcement learning agents leveraging experience replay \citep{hessel_rainbow_2017}.

Among proposed prioritized sampling schemes, \gls{acr:per} remains the most widely used (see Appendix~\ref{ssec:literature-review}). \gls{acr:per} was introduced in~\citet{schaul_prioritized_2015}. It samples transitions in proportion to their absolute \gls{acr:tde}, which measures the distance between predicted and target Q-values. Accordingly, \gls{acr:per} follows the rationale that transitions with higher absolute \glspl{acr:tde} bear higher learning potential. While this rationale is intuitive, the \gls{acr:tde} is a biased proxy as both the predicted and the target Q-value are approximations. Hence, prioritizing transition selection based on absolute \glspl{acr:tde} can misdirect learning, potentially leading to degrading value estimates, if the target Q-value is itself inaccurate \citep{panahi2024investigatinginterplayprioritizedreplay, carrascodavis2025uncertaintyprioritizedexperiencereplay}.
Such inaccurate targets may dampen convergence or in the worst case deteriorate final policy performance.

To address the bias while retaining the efficient transition selection, we propose \gls{acr:reaper}, an enhanced experience replay strategy that extends \gls{acr:per} by weighting the \gls{acr:tde} with a measure of target Q-value reliability. This design is motivated by the observation that, when the agent’s estimation of future states is inaccurate, the corresponding target Q-values become unreliable. In such cases, the \gls{acr:tde} ceases to be a dependable indicator of a transition’s learning potential, leading to ineffective or even detrimental updates. By explicitly accounting for reliability, \gls{acr:reaper} preserves the advantages of \gls{acr:per} over uniform experience replay while mitigating the negative impact of misleading priorities, resulting in consistent performance improvements.



\begin{figure}[!ht]
    \centering
    \includegraphics[width=0.5\linewidth]{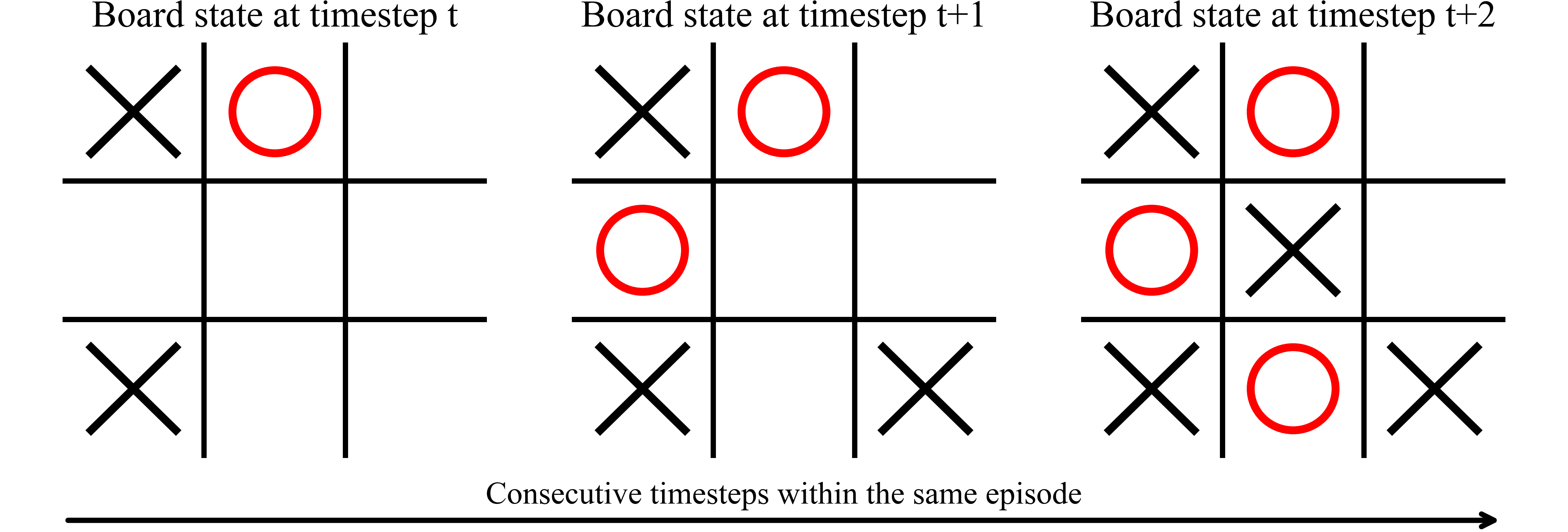}
    \caption{Subsequent states from a Tic Tac Toe game from the perspective of the agent placing circles. Board state $\idxPrimary+2$ is terminal. States $\idxPrimary$ and $\idxPrimary+1$ are losing under optimal play. For an inexperienced player, recognizing that $\idxPrimary+1$ is a losing state is generally easier than recognizing $\idxPrimary$ as such. However, once $\idxPrimary+1$ is understood as losing, identifying $\idxPrimary$ as lost becomes more straightforward. In general, accurately assessing $\idxPrimary+1$ is a prerequisite for reliably assessing $\idxPrimary$—especially when learning the game without explicit knowledge of rules or win conditions. As long as the agent's assessment of $\idxPrimary+1$ is flawed, its evaluation of $\idxPrimary$ remains unreliable.}
    \label{fig:tictactoe}
\end{figure}

\paragraph{Intuition}
The rationale behind our reliability estimate becomes particularly intuitive in game environments such as Go, Chess, or Tic Tac Toe. Consider a player assessing the current board position: if they lack a reliable understanding of how the game might unfold, their evaluation of the current state's value is likely inaccurate. As shown in Figure~\ref{fig:tictactoe}, states closer to terminal outcomes (i.e., near the end of the game) involve fewer remaining moves, 
making it easier for the agent to accurately estimate their values. Early-game states, in contrast, rely on longer and more uncertain rollouts. Thus, value estimates tend to be more reliable as one moves closer to the end of an episode. This observation implies a hierarchical dependency in the learning of transitions within a trajectory, wherein the accurate estimation of earlier state-action values is conditioned on the agent’s ability to infer and propagate information from later transitions. Consequently, we suggest that the reliability of target values -- and by extension, of \glspl{acr:tde} -- should factor into experience replay prioritization.

\paragraph{State of the Art}
Experience replay has been an active field of research for decades. After its first conceptualization by \citet{lin_self-improving_1992}, various extensions, analyses and refinements have been proposed (e.g., \citet{andrychowicz_hindsight_2017, zhang_deeper_2017, isele_selective_2018, rolnick_experience_2018, 
rostami_complementary_2019,fedus_revisiting_2020, lu_synthetic_2023}). 

Central to our work is an active stream of research exploring optimized selection of experiences from the replay buffer. The most notable contribution in this stream so far was \gls{acr:per} \citep{schaul_prioritized_2015, panahi2024investigatinginterplayprioritizedreplay}. 
In essence, \gls{acr:per} proposes to use the absolute \gls{acr:tde} as a sampling weight, which allows to select transitions with high learning potential more frequently compared to a uniform sampling strategy. Various papers built upon the idea of using transition information as a transition selection criterion: \citet{ramicic_entropy-based_2017} explored an entropy-based selection criterion. \citet{gao_prioritized_2021} proposed using experience rewards for sample prioritization. 
\citet{brittain_prioritized_2019} introduced Prioritized Sequence Experience Replay, which extends \gls{acr:per} by propagating absolute \glspl{acr:tde} backwards throughout the episode before using them as a sampling criterion. \citet{zha_experience_2019} and \citet{oh_learning_2021} proposed dynamic, learning-based transition selection mechanisms.  Yet, the proposed approaches have not replaced \gls{acr:per}: \gls{acr:per} remains the only prioritized sampling strategy that is widely adopted by state-of-the-art \gls{acr:rl} algorithms. For a comprehensive review of the relevant literature and a systematic breakdown of this claim, we refer to Appendix~\ref{ssec:literature-review}.

\paragraph{Contribution}
 We propose ReaPER, a novel experience replay sampling scheme that improves upon \gls{acr:per} by reducing the influence of unreliable TD targets, ultimately leading to more stable learning and better policy performance. Specifically, our contribution is threefold: first, we propose the concept of target Q-value and \gls{acr:tde} reliability and introduce a reliability score based on the absolute \glspl{acr:tde} in subsequent states of the same trajectory. Second, we present formal results proving the effectiveness of the reliability-adjusted absolute \gls{acr:tde} as a transition selection criterion. 
Third, we leverage the theoretical insights and the novel reliability score to propose \gls{acr:reaper}, a sampling scheme facilitating more effective experience replay. The proposed method is algorithm-agnostic and can be used within any off-policy \gls{acr:rl} algorithm.

To substantiate our theoretical findings, we perform numerical experiments comparing \gls{acr:reaper} to \gls{acr:per} across various traditional \gls{acr:rl} environments, namely \textsc{CartPole}, \textsc{Acrobot}, \textsc{LunarLander} and the \textsc{Atari-10} benchmark, which recovers 99.2\% of variance within the median score estimate of the full Atari-57 benchmark \citep{aitchison_atari-5_2022}. We show that both prioritized sampling strategies outperform uniform experience replay, and further show that \gls{acr:reaper} consistently outperforms \gls{acr:per}. Specifically, in environments of lower complexity like \textsc{CartPole}, \textsc{Acrobot} and \textsc{LunarLander}, \gls{acr:reaper} reaches the maximum score between $16.6$\% and $32.6$\% faster than \gls{acr:per}. In environments of higher complexity, exemplified by the \textsc{Atari-10} benchmark, \gls{acr:reaper} achieves a $22.97$\% higher median peak performance. In a partially observable variant of the \textsc{Atari-10} benchmark, the performance gap widens, with \gls{acr:reaper} achieving a $34.98$\% median improvement.

\enlargethispage{-0.6cm}

\section{Problem statement}\label{sec:problem_statement}
We consider a standard \gls{acr:mdp} as usually studied in an \gls{acr:rl} setting\citep{sutton_reinforcement_1998}. We characterize this \gls{acr:mdp} as a tuple $\left(\mathcal{S}, \mathcal{A}, P, r, \gamma, p\right)$, where $\mathcal{S}$ is a finite state space, $\mathcal{A}$ is a finite action space, $P: \mathcal{S} \times \mathcal{A} \to \Delta(\mathcal{S})$ is a stochastic kernel, $r: \mathcal{S} \times \mathcal{A} \to \mathbbm{R}$ is a reward function, $\gamma\in(0, 1)$ is a discount factor, and $p\in\Delta(\mathcal{S})$ denotes a probability mass function denoting the distribution of the initial state, $S_1\sim p$. At time step $\idxPrimary$, the system is in state $S_{\idxPrimary}=s\in\mathcal{S}$. We denote by $S_{\idxPrimary}$ and $A_{\idxPrimary}$ the random variables representing the state and action at time $\idxPrimary$, and by $s \in \mathcal{S}$ and $a \in \mathcal{A}$ their respective realizations. If an agent takes action $A_{\idxPrimary}=a\in\mathcal{A}$, it receives a corresponding reward $r(s,a)$, and the system transitions to the next state $S_{\idxPrimary+1}\sim P(\cdot|s,a)$. We define the random reward at time $\idxPrimary$ as $R_{\idxPrimary} = r(S_{\idxPrimary},A_{\idxPrimary})$. The agent selects actions based on a policy $\pi:\mathcal{S}\to\mathcal{A}$ via $A_{\idxPrimary} = \pi(S_{\idxPrimary})$.

Let $\mathbbm{P}_{p}^{\pi}(\cdot)=\textup{Prob}(\cdot\mid\pi,S_1\sim p)$ denote the probability of an event when following a policy $\pi$, starting from an initial state $S_1\sim p$, and let $\mathbbm{E}_p^\pi[\cdot]$ denote the corresponding expectation operator. We consider problems with finite episodes, where $n$ expresses the number of transitions within the episode. Let $G_{\idxPrimary}$ denote the discounted return at time $\idxPrimary$,
$G_{\idxPrimary} = \sum_{\idxSecondary=\idxPrimary}^n \gamma^{\idxSecondary-t}R_{\idxSecondary}$. We define the Q-function (or action-value function) for a policy $\pi$ as

\begin{equation} \label{eq:qfunction}
Q^\pi(s, a) = \mathbbm{E}_p^\pi\left[G_{\idxPrimary} \mid S_{\idxPrimary}=s, A_{\idxPrimary}=a\right]=\mathbbm{E}_p^\pi\left[\sum_{\idxSecondary=\idxPrimary}^n \gamma^{\idxSecondary-\idxPrimary}R_{\idxSecondary} \mid S_{\idxPrimary}=s, A_{\idxPrimary}=a\right].
\end{equation}

The ultimate goal of RL is to learn a policy that maximizes the Q-function, leading to $Q^\star(s,a) = \max_\pi Q^\pi(s,a)$. The policy is gradually improved by repeatedly interacting with the environment and learning from previously experienced transitions. A transition $C_{\idxPrimary}$ is a 5-tuple, $C_{\idxPrimary} = (S_{\idxPrimary}, A_{\idxPrimary}, R_{\idxPrimary}, S_{\idxPrimary+1}, d_{\idxPrimary})$, where $d_{\idxPrimary}$ is a binary episode termination indicator, $d_{\idxPrimary} = \mathbbm{1}_{\idxPrimary = n}$. One popular approach to learn $Q^\star$ is via Watkins' Q-learning \citep{watkins_learning_1989, watkins_q-learning_1992}, where Q-values are gradually updated via
\begin{equation} \label{eq:proposed:update}
Q(S_{\idxPrimary}, A_{\idxPrimary}) \gets Q(S_{\idxPrimary}, A_{\idxPrimary}) + \eta \cdot \delta_{\idxPrimary}
\end{equation}
in which $\eta \in (0,1]$ is the learning rate and $\delta_{\idxPrimary}$ the \gls{acr:tde},
${\delta_{\idxPrimary} = Q_{\text{target}}(S_{\idxPrimary}) - Q(S_{\idxPrimary}, A_{\idxPrimary})}$ with $Q_{\text{target}}(S_{\idxPrimary}) = R_{\idxPrimary+1} + (1 - d_{\idxPrimary+1}) \cdot \gamma \cdot \max_a Q(S_{\idxPrimary+1}, a)$. For brevity of notation, we refer to the absolute \gls{acr:tde} as $\delta_{\idxPrimary}^{+} = \lvert\delta_{\idxPrimary}\rvert$.

In practical RL deployments, when the Q-function is approximated with a neural network, this framework has been augmented with several influential extensions, including \gls{acr:ddqn}, target networks, and experience replay (see Appendix~\ref{ssec:supp-background}). Experience replay is commonly employed to stabilize and accelerate learning. Transitions collected through agent-environment interaction are stored in a finite buffer $\mathcal{H} = \{C_\idxPrimary\}_{\idxPrimary=1}^{N}$, from which mini-batches $\mathcal{X} \subset \mathcal{H}$ of fixed size $\lvert\mathcal{X}\rvert = k$ are sampled to update the Q-function. The sampling distribution over the buffer, denoted by $\mu \in \Delta(\mathcal{H})$, determines the likelihood $\mu(C_\idxPrimary)$ of selecting transition $C_\idxPrimary \in \mathcal{H}$ when constructing $\mathcal{X}$. In uniform experience replay, $\mu$ is the uniform distribution, whereas in \gls{acr:per} \citep{schaul_prioritized_2015}, transitions are sampled according to scalar priority values, derived from the absolute \gls{acr:tde} $\delta_\idxPrimary^+$.

Empirical evidence suggests that the effectiveness of the learning process is sensitive to the choice of $\mu$, i.e., sampling transitions with high learning potential can improve convergence speed and final performance. However, designing an optimal or near-optimal sampling distribution remains an open problem. With this work, we aim to contribute to closing this gap by defining and efficiently approximating a sampling distribution $\mu^\star$ that maximizes learning progress using experience replay.

\section{Methodology}\label{sec:methodology}

In the following, we provide the methodological foundation for \gls{acr:reaper}. We first introduce a reliability score for absolute \glspl{acr:tde}, which we use to derive a \gls{acr:tde}-based reliability-adjusted transition sampling method. We then provide theoretical evidence for its efficacy.

\subsection{Reliability score}\label{sec:reliability:score}
In bootstrapped value estimation, as in Q-learning, the target value
\begin{equation}
Q_{\text{target}}(S_t) = R_{t+1} + \gamma \cdot (1 - d_{t+1}) \cdot \max_a Q(S_{t+1}, a)
\end{equation}
relies on the current estimate of future values. Consequently, the quality of an update to $Q(S_t, A_t)$ depends not only on the magnitude of the absolute \gls{acr:tde} $\delta_t^+$, but also on the reliability of the target value $Q_{\text{target}}(S_t)$.

We define the reliability of a target Q-value as a measure of how well it approximates the true future return from a given state-action pair. 
Intuitively, a target value is reliable if it decreases the distance between $Q^\star(S_t, A_t)$ and $Q(S_t, A_t)$. Conversely, a target is unreliable if training on it increases the distance between $Q^\star(S_t, A_t)$ and $Q(S_t, A_t)$.

To motivate this concept and formalize its operational consequences, we consider a single episode consisting of transitions $(C_1, \dots, C_n)$, from initial state $S_1$ to terminal state $S_{n+1}$. We highlight three key observations that explain how reliability varies along the trajectory and how it can be used to improve sampling:

\begin{observation}[Unreliable targets can degrade learning]
$Q_{\text{target}}(S_t)$ depends on the estimate $Q(S_{t+1}, \cdot)$ for $\idxPrimary \in \{1,\dots,n-1\}$, which may be inaccurate. If an update is based on a poor target value, the resulting $Q(S_t, A_t)$ may diverge from $Q^\star(S_t, A_t)$, thereby degrading the estimate.
\end{observation}

\begin{observation}[Terminal transitions induce reliable updates]
For terminal transitions, the target is given directly by the environment, i.e., $Q_{\text{target}}(S_n) = R_n$. This target is exact, implying that the corresponding \gls{acr:tde} accurately reflects the deviation from the ground truth Q-value. Thus, updates based on terminal transitions are guaranteed to shift $Q(S_n,A_n)$ towards $Q^\star(S_n, A_n)$ if $\delta_n^+ > 0$.
\end{observation}

\begin{observation}[Reliability propagates backward]
An accurate update to $Q(S_t, A_t)$ improves the accuracy of $Q_{\text{target}}(S_{t-1})$ and earlier targets, which recursively depend on it. Therefore, updating transitions near the end of the episode helps improving the reliability of $Q_{\text{target}}$ for earlier transitions.
\end{observation}

 These observations highlight a temporal hierarchy in transition learning: \emph{Learning later transitions before learning earlier transitions appears advantageous}. On the one hand, later targets rely on fewer estimated quantities and are therefore more reliable. On the other hand, learning later transitions positively impacts the target reliability for earlier transitions. Furthermore, a high \gls{acr:tde} indicates a misunderstanding of game dynamics for a given transition, thus rendering the value estimation in predecessor transitions -- which rely on the understanding of the value dynamics in the subsequent rollout -- less reliable. We therefore aim to resolve \glspl{acr:tde} back-to-front. This motivates defining the reliability of $Q_{\text{target}}(S_t)$ inversely related to the sum of future absolute \glspl{acr:tde}, 

\begin{equation} \label{eq:reliability}
    \mathcal{R}_t = 1 - \frac{\sum_{\idxSecondary = t+1}^{n} \delta_\idxSecondary^+}{\sum_{\idxSecondary = 1}^{n} {\delta_{\idxSecondary}^+}}.
\end{equation}
Using this definition, we propose the \emph{reliability-adjusted \gls{acr:tde}}
\begin{equation} \label{eq:reliability_adjusted_tde}
    \Psi_t = \mathcal{R}_t \cdot \delta_t^+,
\end{equation}
as a sampling criterion for selecting transitions during experience replay. High values of $\Psi_t$ correspond to transitions that promise large updates and have reliable target values. Sampling weights $p$ can be obtained by normalizing the sampling criterion with the sum of $\Psi$ over all transitions.

\subsection{Formal analysis}\label{ssec:formal_analysis}
We consider a set of transitions that constitutes a single complete trajectory of length $n$, \( \mathcal{D} = \{C_t\}_{t=1}^n \), where \( C_t = (S_t, A_t, R_t, S_{t+1}, d_t) \).

Updates are based on the \gls{acr:tde} \( \delta_t = Q(S_t, A_t) - Q_{\text{target}}(S_t) \), using the standard bootstrapped target
\begin{equation}
Q_{\text{target}}(S_t) = R_t + \gamma (1 - d_t) \max_a Q(S_{t+1}, a).
\end{equation}

\paragraph{Convergence} A critical factor to ensure convergence in Q-learning is the alignment between the \gls{acr:tde} and the true value estimation error $Q(S_t, A_t) - Q^\star(S_t, A_t)$. When the bootstrapped target is biased, meaning \( Q_{\text{target}}(S_t) \neq Q^\star(S_t, A_t) \), the update direction may become misaligned, potentially worsening the value estimate.

We defer the formal misalignment analysis to Lemma~\ref{lemma:misalignment} and Lemma ~\ref{lemma:update_expectation} in Appendix~\ref{sssec:convergence_behavior}. In essence, the expected change in squared true value estimation error due to an update of the value function approximator under a sampling strategy \( \mu \) can be decomposed into three components, the \gls{acr:tde} variance, the true squared error, and the bias-error interaction
\begin{equation}
\mathbbm{E}_{\mu}[\Delta \lvert Q(S_t, A_t) - Q^\star(S_t, A_t)\rvert^2] = \underbracket[0.5pt]{\eta^2 \sum_{t=1}^{n} \mu_t \mathbbm{E}[\delta_t^2]}_{\text{\gls{acr:tde} variance}} - \underbracket[0.5pt]{2\eta \sum_{t=1}^{n} \mu_t \mathbbm{E}[e_t^2]}_{\text{True squared error}} + \underbracket[0.5pt]{2\eta \sum_{t=1}^{n} \mu_t \mathbbm{E}[e_t\varepsilon_t]}_{\text{Bias-error-interaction}} ,
\end{equation}

where $e_t$ denotes the true value error, and $\varepsilon_t$ denotes the target bias, 

\begin{equation}
    e_t = Q(S_t, A_t) - Q^\star(S_t, A_t), \quad \varepsilon_t = Q_{\text{target}}(S_t) - Q^\star(S_t, A_t).
\end{equation}

By focusing on large \glspl{acr:tde}, PER aims to sample transitions with higher true squared error more frequently, thus resolving errors faster and improving efficiency over uniform sampling. In the following, we show that our ReaPER sampling scheme additionally controls the target bias, thereby minimizing the bias-error interaction, while also preserving the advantages of \gls{acr:per}. This allows \gls{acr:reaper} to increase sampling-efficiency over PER. To do so, we base the following formal analyses on a key assumption relating target bias to absolute downstream \glspl{acr:tde}.

\begin{assumption}[Target Bias via Downstream \glspl{acr:tde}]
\label{assumption:bias}
Along an optimal trajectory, the target bias $\varepsilon_t$ for each transition $C_t$ satisfies
\begin{equation}
\lvert\varepsilon_t\rvert \leq \lambda \sum_{i = t+1}^{n} \delta_i^+,
\end{equation}
\end{assumption} 
for some \( \lambda \leq 1 \), where \( \delta_i^+ = \lvert\delta_i\rvert \). This assumption formalizes the intuition that bootstrapped targets primarily inherit bias from inaccuracies in future predictions. It reflects standard TD-learning dynamics under sufficient exploration and function approximation stability. While -- similar to the assumptions made in standard convergence proofs for \gls{acr:rl} -- this assumption may theoretically be violated during the early phases of training, it tends to hold in practice, especially once value estimates stabilize, as we show in Appendix~\ref{sec:assumption_testing}. In a nutshell, Assumption~\ref{assumption:bias} captures the intuition that target bias predominantly arises from unresolved downstream \glspl{acr:tde}, reflecting a local perspective on TD-learning dynamics. Unlike classical convergence proofs that require global exploration and decaying learning rates, our assumption focuses on bounding the bias along observed trajectories during finite-sample learning, making it more applicable to practical deep \gls{acr:rl}. We refer the interested reader to Appendix~\ref{sssec:assumption} for a detailed discussion.

Under Assumption~\ref{assumption:bias}, the reliability score \( \mathcal{R}_t \) --representing the fraction of downstream \gls{acr:tde} within a given trajectory -- bounds the normalized target bias. This is captured in the following lemma.

\begin{lemma}[Reliability Bounds Target Bias]
\label{lemma:reliability}
Under Assumption~\ref{assumption:bias},
\begin{equation}
\lvert\varepsilon_t\rvert \leq \lambda (1 - \mathcal{R}_t) \sum_{i=1}^n \delta_i^+.
\end{equation}
\end{lemma}

For the proof of Lemma~\ref{lemma:reliability}, we refer to Appendix~\ref{sssec:reliability_bias_bound}. Lemma~\ref{lemma:reliability} expresses that transitions with higher reliability scores exhibit lower target bias and thus yield more trustworthy \glspl{acr:tde}. This finding motivates selecting training transitions not just by \gls{acr:tde} magnitude, but by a combination of \gls{acr:tde} magnitude and reliability — as implemented in \gls{acr:reaper}.

Building on the established relationship between reliability and target bias, we derive the following convergence hierarchy.

\begin{proposition}[Convergence Hierarchy of Sampling Strategies]
\label{prop:hierarchy}
Under Assumption~\ref{assumption:bias} and given a fixed learning rate \( \eta \), \gls{acr:reaper} (\( \mu_t \propto \mathcal{R}_t \delta_t^+ \)) yields lower expected Q-value error than standard \gls{acr:per} (\( \mu_t \propto \delta_t^+ \)), which in turn outperforms uniform sampling,
\begin{equation}
\mathbbm{E}[|| Q_T^{\mathrm{(Uniform)}} - Q^\star||^2]
\geq
\mathbbm{E}[|| Q_T^{\mathrm{(PER)}} - Q^\star||^2]
\geq
\mathbbm{E}[|| Q_T^{\mathrm{(ReaPER)}} - Q^\star|| ^2],
\end{equation}
where $\mathbbm{E}$ denotes the expectation across training runs.
\end{proposition}

The corresponding proof, detailed in Appendix~\ref{sssec:hierarchy}, formally compares the expected error decrease terms under different sampling distributions, using Lemma~\ref{lemma:reliability} to bound the bias-error-interaction. While we limit Proposition~\ref{prop:hierarchy} to optimal policies for brevity of notation, we can straightforwardly extend it to suboptimal policies.

\begin{remark}[Extension to suboptimal policies]
\label{remark:suboptimal}
If the agent follows a fixed but suboptimal policy, Assumption~\ref{assumption:bias} can be relaxed to include an additive policy-induced bias term \( \zeta \geq 0 \), yielding
\begin{equation}
|\varepsilon_t| \leq \lambda \sum_{i = t+1}^n \delta_i^+ + \zeta.
\end{equation}
In this case, \gls{acr:reaper} still improves sampling efficiency in expectation, although the achievable Q-value accuracy is lower-bounded by the policy suboptimality \( \zeta \). For further details, we refer the interested reader to Appendix~\ref{sssec:remark}.
\end{remark}

Together, these results provide the theoretical foundation for \gls{acr:reaper}'s design: By prioritizing transitions with high absolute \gls{acr:tde} and high reliability, \gls{acr:reaper} selects relevant transitions while improving alignment with the true value error, leading to faster and more stable learning.

\paragraph{Variance reduction} 

In the following, we show that the proposed sampling scheme, based on reliability-adjusted \glspl{acr:tde} reduces the variance of the Q-function updates. As a first step towards this result, we analyze the theoretically optimal distribution to sample from in order to minimize the variance of the Q-function update step.
Recall that Q-values are updated according to \eqref{eq:proposed:update}, where the \gls{acr:tde} corresponding to a transition $C_t$ from the finite replay buffer $\mathcal{H}$ reads $\delta_t = Q_{\text{target}}(S_t) - Q(S_t, A_t)$. 

We assume a fixed episode and treat the current $Q$-values as constants, focusing on analyzing the update variance induced by the sampling distribution $\mu$ over $\mathcal{H}$.
The update variance can then be expressed as
\begin{equation}\label{eq:update:variance}
    \sum_{t=1}^N \mu_t \mathsf{Var}[\delta_t] = \sum_{t=1}^N \mu_t \mathsf{Var}[Q_\text{target}(S_t)] = \sum_{t=1}^N \mu_t \sigma_t^2,
\end{equation}
where the first equality follows from the definition of the \gls{acr:tde} and the assumption that the current $Q$-values are constant. The second equality simply defines $\sigma_i^2 := \mathsf{Var}[Q_\text{target}(S_i)]$ as the variance of the bootstrapped target for brevity.

\begin{proposition}[Variance reduction via reliability-aware sampling]\label{prop:variance:reduction}
The distribution $\mu^\star$ minimizing the update variance \eqref{eq:update:variance} is given by
\begin{equation}
    \mu^\star_t \propto \frac{\delta_t^+}{\sigma_t^2} \text{ for all } t \in \{1,\dots,N\}
\end{equation}
\end{proposition}

For the proof of Proposition~\ref{prop:variance:reduction}, we refer to Appendix~\ref{sssec:variance_reduction}.
 
As a direct consequence of Proposition~\ref{prop:variance:reduction}, we find that our proposed \gls{acr:reaper} sampling scheme is variance reducing if the reliability $\mathcal{R}$ is proportional to the inverse variance of the bootstrapped target. 

As the true target $Q^\star$ remains constant, a significant proportion of variance across runs for a given state can be attributed to the target bias. As such, there exists a direct relationship between $\varepsilon$ and $\sigma^2$. Hence, under Assumption~\ref{assumption:bias}, it seems natural to assume $\mathcal{R} \propto \frac{1}{\sigma^2}$. Thus, \gls{acr:reaper} constitutes a reasonable proxy for the optimal inverse-variance weighted sampling strategy.

To provide further intuition for our formal results, we have conducted a supplementary simulation-driven analysis showing that \gls{acr:reaper} achieves optimal transition selection in a stylized setting, which we detail in Appendix~\ref{ssec:stylized_setting}.

\section{Reliability-adjusted Prioritized Experience Replay} \label{sec:reliability_adjusted_prioritized_experience_replay}

We have thus far introduced the reliability-adjusted \gls{acr:tde} and theoretically proven its effectiveness as a transition selection criterion. In the following, we propose \gls{acr:reaper}, the sampling algorithm built around the reliability-adjusted \gls{acr:tde}. We give a distilled overview of the resulting sampling scheme in Algorithm~\ref{algo:reaper}. At its core, we create mini-batches by sampling from the buffer with $\Psi$ as the sampling weight. Specifically, at each training step $\tau \in \{1, \dots, T\}$, for every transition within the buffer, that is, $C_t$ for all $t \in \{1,\dots,N\}$, we update \gls{acr:tde} reliabilities $\mathcal{R}_t$ and compute the transition selection criterion $\Psi_t$ (Algorithm~\ref{algo:reaper}, Line~\ref{algline:update_loop}ff.). Based on $\Psi_t$, we sample $k$ transitions from the buffer $\mathcal{H}$ to create the next mini-batch $\mathcal{X}$ (Algorithm~\ref{algo:reaper}, Line~\ref{algline:sample_loop}ff.). For the full algorithm and an extended explanation, we refer to Appendix~\ref{ssec:algorithm}.

\vspace{-0.3pt}
\begin{algorithm}[h]
\caption{Sampling transitions and updating the value function using \gls{acr:reaper}}\label{algo:reaper}
\KwIn{absolute \glspl{acr:tde} $\delta^+$, episode vector $\phi$, current episode $\Phi$, batch size $k$, exponents $\alpha$, $\omega$ and $\beta$, replay buffer $\mathcal{H}$ of size $N$, policy weights $\theta$, maximum priority $p_{max}$ = 1, budget $T$}

\For{$\tau \in \{1, \dots, T\}$}{
    Initialize accumulated weight change $\Delta = 0$ and empty batch $\mathcal{X} = \bm{0}^{(k)}$\;
    
    Add novel transitions to the buffer with maximum priority $p_{max}$ and set $\phi_t = \Phi$\;
    
    Compute maximum episodic sum of absolute \glspl{acr:tde}, $F \gets \max\limits_{\idxPrimary\in\{1,\dots,N\}} \left( \sum_{\idxSecondary=1}^{N} \delta_{\idxSecondary}^+ \cdot \mathbbm{1}_{\phi_{\idxPrimary} = \phi_i} \right)$\;
    
    \For(\tcp*[f]{Updating transition weights}){$\idxPrimary \in \{1, \dots, N\}$}{ \label{algline:update_loop}
        Compute \gls{acr:tde} reliabilities as in Formula~\ref{reliability_final}\; 
        Compute transition selection criterion $\Psi_\idxPrimary \gets \mathcal{R}_\idxPrimary^\omega \cdot {\delta_\idxPrimary^+}^\alpha$\;
        Compute transition priorities $p_\idxPrimary \gets \frac{\Psi_\idxPrimary}{\sum^{N}_{\idxSecondary=1} \Psi_{\idxSecondary}}$\;
    }
    \For(\tcp*[f]{Sampling transitions}){$\idxTertiary \in \{1, \dots, k\}$}{ \label{algline:sample_loop}
        Sample a transition $C_j$ from $\mathcal{H}$ to add to batch $\mathcal{X}$ such that $\mathbbm{P}(C_\idxPrimary=\mathcal{X}_\idxTertiary) = p_\idxPrimary$ for all $\idxPrimary \in \{1, \dots, N\}$\;
        Compute importance-sampling weight $w_{\idxSelected} \gets \frac{\left(N \cdot p_\idxSelected\right)^{-\beta}}{\max_{\idxPrimary} w_{\idxPrimary}}$ for all $\idxPrimary \in \{1, \dots, N\}$\;
        Update $\delta_j^+$ and accumulate weight-change $\Delta \gets \Delta + w_{\idxSelected} \cdot \delta_{\idxSelected} \cdot \nabla_\theta Q(S_{\idxSelected}, A_{\idxSelected})$\;

    }
    
    Update weights $\theta \gets \theta + \eta \cdot \Delta$\;
    
    Update maximum priority $p_{max} = \max(p_{max}, \max(p))$\;
}
\end{algorithm}
\vspace{-0.5pt}

Starting from the naive implementation of \gls{acr:reaper}, we require four technical refinements to obtain a functional and efficient sampling algorithm.

I. \emph{Priority updates.} To consistently maintain an updated sampling weight $\Psi$, we track the \glspl{acr:tde} and reliabilities of stored transitions throughout the training. As it is computationally intractable to re-calculate all \glspl{acr:tde} on every model update, we implement a leaner update rule: As in \gls{acr:per}, we assign transitions maximum priority when they are added to the buffer. Moreover, we assign the \gls{acr:tde} of transition $C_\idxPrimary$ every time $C_\idxPrimary$ is used to update the Q-function. We assign the reliability of transition $C_t$ every time $C_t$ is used to update the Q-function, or if any other transition from the same episode is used to update the Q-function, as it leads to a change in the sum of \glspl{acr:tde} and possibly the subsequent \glspl{acr:tde}. We update the priority $\Psi$ when \gls{acr:tde} or reliability are updated.

II. \emph{Priority regularization.} As the \gls{acr:tde} of a given transition may change by updating the model even without training on this transition, \glspl{acr:tde} -- and, in consequence, reliabilities -- are not guaranteed to be up-to-date. Thus, similar to \citet{schaul_prioritized_2015}, we introduce regularization exponents $\alpha \in (0,1]$ and $\omega \in (0,1]$ to dampen the impact of extremely high or low \glspl{acr:tde} or reliabilities,
\begin{equation}\label{eq:sampling_proba}
  \Psi_\idxPrimary = \mathcal{R}_\idxPrimary^\omega \cdot {\delta_\idxPrimary^+}^\alpha.
\end{equation}
III. \emph{Reliabilities for ongoing episodes.} As the sum of \gls{acr:tde} throughout an episode is undefined as long as the episode is not terminated, so is the reliability. In these cases, we use the maximum sum of \glspl{acr:tde} of any episode within the buffer to obtain a conservative reliability estimate.

For this, we introduce $\phi$, a vector of length $N$, where $\phi_t$ denotes the $t$-th position in $\phi$, which contains a scalar counter of the trajectory during which transition $C_t$ was observed. As such, $\phi$ functions as a positional encoding of transitions within the buffer. Specifically, it is used to identify all transitions that belong to the same trajectory. This positional encoding allows us to calculate conservative reliability estimates for a multi-episodic buffer.

We then define $\mathcal{R}_t$ as

\begin{equation}\label{reliability_final}
    \mathcal{R}_t = 
    \begin{cases}
        1 - \left(\frac{\sum_{i = t+1}^{n} \delta_i^+}{\sum_{i = 1}^{n} \delta_i^+} \right)& \text{for transitions of terminated episodes} \\
        1 - \left( \frac{F - \sum_{i = 1}^{t} \delta_i^+}{F} \right) & \text{for transitions of ongoing episodes}
    \end{cases}
\end{equation}

where 

\begin{equation}
F = \max\limits_{\idxPrimary\in\{1,\dots,N\}} \left( \sum_{\idxSecondary=1}^{N} \delta_{\idxSecondary}^+ \cdot \mathbbm{1}_{\phi_{\idxPrimary} = \phi_i} \right)
\end{equation}.

Importantly, this formulation of reliability focuses on within-episodic variance instead of inter-episodic comparability. It therefore deliberately does not account for differences in episode length, as this introduces bias in favor of shorter trajectories, which we found to be detrimental in practice. 

VI) \emph{Weighted importance sampling.} Finally, just as every other non-uniform sampling method, \gls{acr:reaper} violates the i.i.d. assumption. Thus, it introduces bias into the learning process, which can be harmful when used in conjunction with state-of-the-art \gls{acr:rl} algorithms. Similar to \citet{schaul_prioritized_2015}, we use weighted importance sampling \citep{mahmood_off-policy_2014} to mitigate this bias. When using importance sampling, each transition $C_{\idxPrimary}$ is assigned a weight $w_{\idxPrimary}$, such that 

\begin{equation}\label{eq:importance_sampling_weight}
    w_{\idxPrimary} = \left( \frac{1}{N} \cdot \frac{1}{p_\idxPrimary} \right)^\beta = \left( \frac{1}{N} \cdot \frac{\sum_{i=1}^N\Psi_i}{\Psi_\idxPrimary} \right)^\beta.
\end{equation}

We use this weight to scale the loss and perform Q-learning updates using $\delta_{\idxPrimary} \cdot w_{\idxPrimary}$ instead of~$\delta_{\idxPrimary}$.

\section{Numerical study}\label{sec:experiments}

We evaluated \gls{acr:reaper} against \gls{acr:per} across a diverse set of continuous control and Atari environments. For continuous control, we considered the discrete action space environments from the Gymnasium library \citep{towers_gymnasium_2024}, namely \textsc{CartPole}, \textsc{Acrobot} and \textsc{LunarLander}. For Atari, following prior work, we use \textsc{Atari-10} as a computationally efficient yet representative benchmark which recovers 99.2\% of median score variance within the Atari-57 benchmark, ensuring relevance to broader Atari-57 evaluations without incurring prohibitive computational overhead \citep{aitchison_atari-5_2022}. Across conditions, we used the same \gls{acr:ddqn} agent, neural architecture, and model hyperparameters for all experiments. We controlled for all sources of randomness using fixed seeds and compared algorithms using identical seeds per trial. Thus, the only variation between conditions stemmed from the experience replay algorithm. Full experimental details and hyperparameters are provided in Appendix~\ref{ssec:experimental_settings}. For a detailed introduction to additional deep Q-learning–related concepts, we refer to Appendix~\ref{ssec:supp-background}.

\paragraph{Continuous control}

For each continuous control environment, we compared the performance between \gls{acr:per} and \gls{acr:reaper} across $20$ training runs. Training ended preemptively when a pre-defined score threshold was met \citep{towers_gymnasium_2024}.

Across all three environments, \gls{acr:reaper} consistently reached performance thresholds in fewer steps than both uniform replay and \gls{acr:per}. In \textsc{Acrobot}, this corresponded to improvements of $25.0\%$ and $16.6\%$, respectively. For \textsc{CartPole}, \gls{acr:reaper} reduced the steps needed by $41.4\%$ and $32.6\%$, and a similar pattern held in \textsc{LunarLander}, with gains of $37.1\%$ and $21.1\%$.

\begin{figure}[!ht]
    \centering
    \includegraphics[width=1\linewidth]{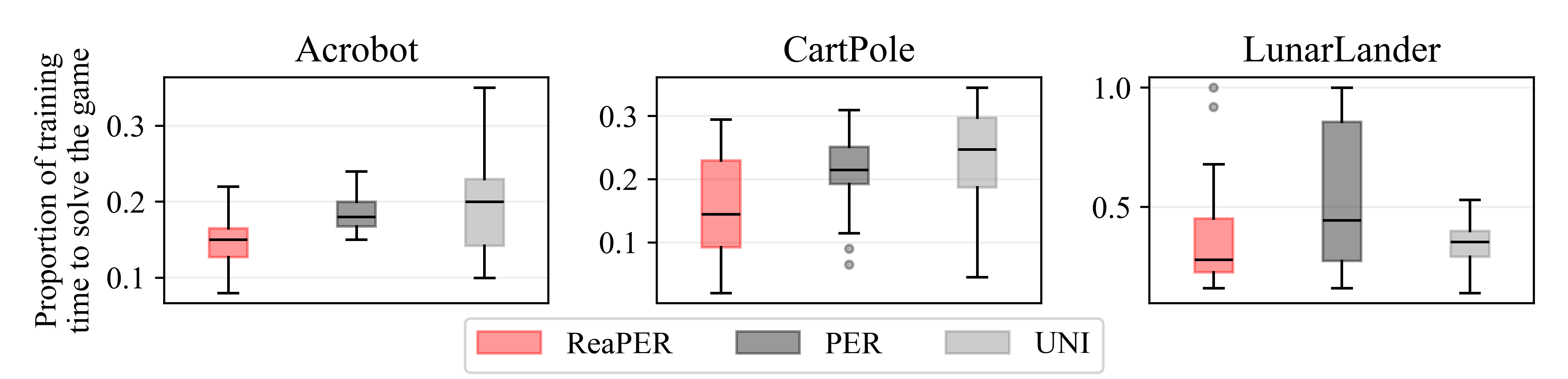}
    \vspace{-0.5cm}
    \caption{Proportion of training steps required by \gls{acr:per}, uniform experience replay (UNI) and \gls{acr:reaper} to reach a pre-defined score thresholds given in \citet{towers_gymnasium_2024} across 20 runs in three traditional \gls{acr:rl} environments. The shaded region corresponds to the interquartile range (IQR), whiskers extend to $\text{Q1} - 1.5 \cdot \text{IQR}$ and $\text{Q3} + 1.5 \cdot \text{IQR}$, and the horizontal bar indicates the median score.}
    \vspace{-0.5cm}
    \label{fig:small_env}
\end{figure}

\begin{figure}[!ht]
    \centering
    \includegraphics[width=1\linewidth]{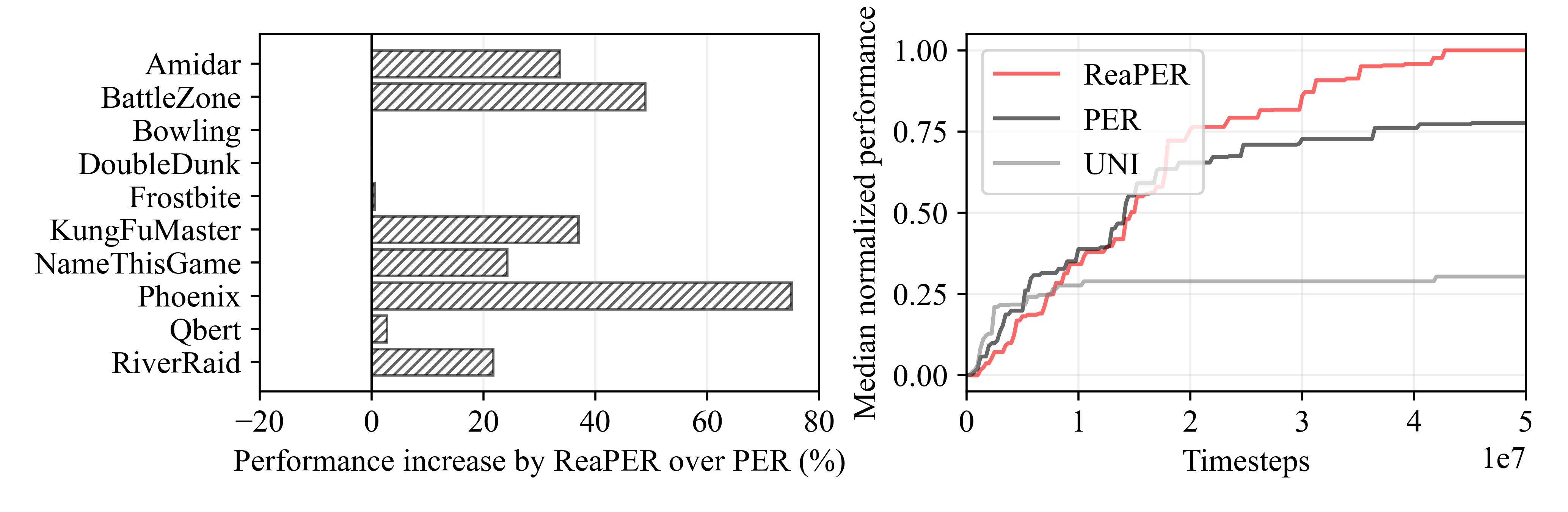}
    \vspace{-0.5cm}
    \caption{Left: Peak score increase of \gls{acr:reaper} over \gls{acr:per}. Right: Median of the normalized cumulative maximum of scores across the Atari-10 benchmark for \gls{acr:reaper}, \gls{acr:per} and uniform experience replay (UNI), following reporting standards from \citet{schaul_prioritized_2015}. The normalized score at timestep $\idxPrimary$ is calculated by dividing the difference between the current score and the random score by the difference between the maximum score in this game across all sampling strategies and the random score.}
    \vspace{-0.5cm}
    \label{fig:atari}
\end{figure}

\paragraph{Atari}

\gls{acr:reaper} consistently outperformed both uniform experience replay and \gls{acr:per} on the \textsc{Atari-10} benchmark. Specifically, \gls{acr:reaper} outperformed \gls{acr:per} and uniform experience replay in eight out of ten games, tying \gls{acr:per} in two games. Across all games, \gls{acr:reaper} achieved a a $22.97$\% higher median peak score than \gls{acr:per}, and a $229.78$\% higher median peak score than uniform experience replay. Under partial observability, \gls{acr:reaper}'s median outperformance grows to $34.98\%$. These results underline \gls{acr:reaper}'s robustness across heterogeneous game dynamics and its ability to scale to challenging, high-dimensional domains. We provide per-game curves in Appendix~\ref{ssec:atari10_results} (Figure~\ref{fig:allatari_cummax}), and extended information on results under partial observability in Appendix~\ref{ssec:partial_observability}.

\paragraph{Discussion} \label{sec:discussion}
\gls{acr:reaper} consistently outperforms \gls{acr:per}, indicating a substantial methodological advance. Notably, \gls{acr:reaper} did so with minimal hyperparameter tuning. We expect further gains through more extensive tuning of key hyperparameters, including regularization exponents $\alpha$ and $\omega$, importance sampling exponent $\beta$ and learning rate $\eta$.

A limitation of \gls{acr:reaper} is its reliance on terminal states, which are a pre-requisite for calculating meaningful \gls{acr:tde} reliabilities. Further, \gls{acr:reaper} tracks the episodic cumulative sums of \glspl{acr:tde} to calculate the reliability score, which causes computational overhead when \glspl{acr:tde} are updated. Using a naive implementation, this overhead is non-negligible at $O(N)$. However, it can be reduced to $O(n-\idxPrimary)$ by only re-calculating the episodic cumulative sums for transitions on their update or the update of a preceding transition within the same episode.


\section{Conclusion}\label{sec:conclusion}
We introduced \gls{acr:reaper}, a reliability-adjusted experience replay method that mitigates the detrimental effects of unreliable targets in off-policy deep reinforcement learning. By formally linking target bias to downstream temporal difference errors, we proposed a principled reliability score that enables more efficient and stable sampling. Our theoretical analysis shows that \gls{acr:reaper} improves both convergence speed and variance reduction over standard \gls{acr:per}, and our empirical results confirm its effectiveness across diverse benchmarks.

Beyond its immediate practical gains, \gls{acr:reaper} highlights the importance of accounting for target reliability in experience replay, particularly in deep \gls{acr:rl} settings where function approximation errors and generalization artifacts are prevalent. We believe our work opens new avenues for incorporating uncertainty and reliability estimates into replay buffers, and future research may explore adaptive reliability estimation, extensions to actor-critic methods and infinite-horizon settings, as well as integration with representation learning.

\newpage

\bibliographystyle{unsrtnat}  
\bibliography{references}  

\newpage

\appendix

\section{Algorithm}\label{ssec:algorithm}

\begin{algorithm}
    \caption{Deep Q Learning with reliability-adjusted proportional prioritization}\label{algo:reaper_ddqn}
    \SetAlgoLined
    \KwIn{batch size $k$, learning rate $\eta$, replay period $K$, replay buffer size $N$, exponents $\alpha$, $\omega$ and $\beta$, budget $T$.}
    Initialize replay memory $\mathcal{H} = \emptyset$, $\Delta = 0$, $p_1 = 1$, episode vector $\phi = \mathbf{0}^{(N)}$, episodic count $\Phi = 1$ and maximum sum of episodic \gls{acr:tde} $F= 1$\;  

    Observe $S_1$ and choose $A_1 \sim \pi_\theta(S_1)$\; \label{alg2line:init_action}
    
    \For{$\idxCurrent \in \{1, \dots, T\}$}{

        Initialize accumulated weight change $\Delta = 0$ and empty batch $\mathcal{X} = \bm{0}^{(k)}$; 
        
            Observe $S_{\idxCurrent+1}$, $R_{\idxCurrent}$, $d_{\idxCurrent}$\; \label{alg2line:observe_transition}
            
            Store transition $C_{\idxCurrent} = (S_{\idxCurrent}, A_{\idxCurrent}, R_{\idxCurrent}, d_{\idxCurrent}, S_{\idxCurrent+1})$ in $\mathcal{H}$ with $\phi_{\idxCurrent} = \Phi$ and $p_{\idxCurrent} = \max_{\idxPrimary}(p_{\idxPrimary})$ for all $\idxPrimary \in \{1, \dots, N\}$\; \label{alg2line:store_transition}
            
            \If{$c \equiv 0 \mod K$}{\label{alg2line:check_training_step}
            
                \For{$\idxTertiary \in \{1, \dots, k\}$}{ \label{alg2line:sample}
                
                    Sample a transition $C_j$ from $\mathcal{H}$ to add to batch $\mathcal{X}$ such that $\mathbbm{P}(C_\idxPrimary=\mathcal{X}_\idxTertiary)  = p_\idxPrimary$ for all $\idxPrimary \in \{1, \dots, N\}$\;
                    
                    Compute importance-sampling weight $w_{\idxSelected} = \frac{\left(N \cdot p_\idxSelected\right)^{-\beta}}{\max_{\idxPrimary} w_{\idxPrimary}}$ for all $\idxPrimary \in \{1, \dots, N\}$\;
                    
                    Compute \gls{acr:tde} $\delta_{\idxSelected} = Q_{target}(S_{\idxSelected}) - Q(S_{\idxSelected}, A_{\idxSelected})$\;
                    
                    Accumulate weight-change $\Delta \gets \Delta + w_{\idxSelected} \cdot \delta_{\idxSelected} \cdot \nabla_\theta Q(S_{\idxSelected}, A_{\idxSelected})$\;
                }

                Update weights $\theta \gets \theta + \eta \cdot \Delta$\; \label{alg2line:update_model}
            
                From time to time, copy weights into target network, $\theta_{\text{target}} \gets \theta$\;

                Update maximum sum of absolute \glspl{acr:tde}, $F \gets \max\limits_{\idxPrimary\in\{1,\dots,N\}} \left( \sum_{\idxSecondary=1}^{N} \delta_{\idxSecondary}^+ \cdot \mathbbm{1}_{\phi_{\idxPrimary} = \phi_i} \right)$\;
                
                \For{$\idxPrimary \in \{1, \dots, N\}$}{ \label{alg2line:update_priority_loop}

                    Compute \gls{acr:tde} reliabilities, $\mathcal{R}_t = 
                        \begin{cases}
                            1 - \left(\frac{\sum_{i = t+1}^{n} \delta_i^+}{\sum_{i = 1}^{n} \delta_i^+} \right)& \text{for transitions of terminated episodes} \\
                            1 - \left( \frac{F - \sum_{i = 1}^{t} \delta_i^+}{F} \right) & \text{for transitions of ongoing episodes}
                        \end{cases}$ \label{alg2line:update_reliabilities}
                    
                    Update transition sampling criterion $\Psi_\idxPrimary \gets \mathcal{R}_\idxPrimary^\omega \cdot {\delta_\idxPrimary^+}^\alpha$\;
                            
                    Update transition priorities $p_\idxPrimary \gets \frac{\Psi_\idxPrimary}{\sum^{N}_{\idxSecondary=1} \Psi_{\idxSecondary}}$\; \label{alg2line:update_priorities}
                }
        }
    
        \If{$d_{\idxCurrent} = 1$}{
            \For{$\idxPrimary \in \{1, \dots, N\} \mid \phi_{\idxPrimary} = \phi_c)$}{
                Compute \gls{acr:tde} reliabilities for the finished episode,  $R_t = 1 - \left(\frac{\sum_{i = t+1}^{n} \delta_i^+}{\sum_{i = 1}^{n} \delta_i^+} \right)$ \;\label{alg2line:compute_reliability_finished_runs}
            }
            $\Phi \gets \Phi + 1$\; \label{alg2line:track_episodes}
        }

        Choose action $A_{\idxCurrent} \sim \pi_\theta(S_{\idxCurrent})$\; \label{alg2line:choose_action}

    }    

\end{algorithm}

In the following, we describe how \gls{acr:reaper} operates in conjunction with a \gls{acr:dqn}. The agent begins by observing the initial state and selecting an action (Algorithm~\ref{algo:reaper_ddqn}, Line~\ref{alg2line:init_action}).

For a fixed number of iterations, the agent interacts with the environment, observes the resulting transition from its latest action, and stores this transition in the replay buffer with maximum priority (Algorithm~\ref{algo:reaper_ddqn}, Line~\ref{alg2line:observe_transition}f.).

Every $K$ steps, the agent performs a training update (Algorithm~\ref{algo:reaper_ddqn}, Line~\ref{alg2line:check_training_step}). During training, it samples a batch $\mathcal{X}$ from the buffer using the current priorities $p$ as sampling weights (Algorithm~\ref{algo:reaper_ddqn}, Line~\ref{alg2line:sample}ff.). The agent updates the model parameters using importance-sampling-weighted TD-errors (Algorithm~\ref{algo:reaper_ddqn}, Line~\ref{alg2line:update_model}), and uses the observed TD-errors to update the priorities $p$ for all transitions in the batch (Algorithm~\ref{algo:reaper_ddqn}, Line~\ref{alg2line:update_priority_loop}ff.). This involves recomputing the reliabilities based on the new TD-errors, applying a conservative estimate for transitions from ongoing episodes (Algorithm~\ref{algo:reaper_ddqn}, Line~\ref{alg2line:update_reliabilities}). The agent then recalculates the sampling criterion $\Phi$ and updates the priorities $p$ accordingly (Algorithm~\ref{algo:reaper_ddqn}, Line~\ref{alg2line:update_priorities}), concluding the training step.

Upon episode termination, the agent replaces the preliminary reliability estimate with the actual reliability (Algorithm~\ref{algo:reaper_ddqn}, Line~\ref{alg2line:compute_reliability_finished_runs}). Throughout training, it tracks episode progress to enable continuous recomputation of reliabilities (Algorithm~\ref{algo:reaper_ddqn}, Line~\ref{alg2line:track_episodes}).

At each iteration, the agent selects the next action based on its current policy and state (Algorithm~\ref{algo:reaper_ddqn}, Line~\ref{alg2line:choose_action}), initiating the next cycle.

\section{Supplementary Background}\label{ssec:supp-background}

\paragraph{Target networks.}
Target networks stabilize \gls{acr:dqn} by maintaining a separate copy of the Q-network whose parameters are updated more slowly. This reduces harmful feedback loops between rapidly changing estimates and improves training stability. In practice, the target network parameters are either periodically copied from the online network or updated via a soft update rule, where parameters slowly track the online network through Polyak averaging. These techniques help ensure that the bootstrap targets change smoothly over time, making optimization substantially more robust. For further information, see \citet{mnih_human-level_2015}.

\paragraph{\gls{acr:ddqn}.}
\gls{acr:ddqn} mitigates Q-value overestimation by decoupling action selection from value evaluation: the online network selects the action, while the target network evaluates it. This leads to more accurate value estimates and often improves policy quality. In addition to reducing positive bias, DDQN can yield more stable learning dynamics, particularly in environments with noisy rewards or large action spaces, where overestimation errors tend to accumulate. For further details, see \citet{van_hasselt_deep_2015}.

\paragraph{Optimizers.}
We use standard adaptive gradient methods. Adam \citep{kingma_adam_2014} maintains per-parameter first- and second-moment estimates to provide smooth, scaled updates. RMSProp adapts learning rates based on an exponentially weighted average of recent squared gradients, helping control step sizes in non-stationary settings. In practice, these methods reduce the sensitivity to hyperparameter choices such as the initial learning rate and improve convergence speed, especially in deep reinforcement learning where gradient magnitudes can vary significantly across parameters and time. We refer to \citet{ruder2017overviewgradientdescentoptimization} for a comprehensive introduction.

\section{Literature Review}\label{ssec:literature-review}

In the following, we substantiate our claim that \gls{acr:per} constitutes the most practically relevant prioritized sampling strategy within reinforcement learning to this day. We first systematically review state-of-the-art \gls{acr:rl} algorithms and the sampling strategies they are employing. We further discuss possible reasons for the limited adoption of proposed alternatives.

\subsection{Prioritized Experience Replay as the state-of-the-art}

When we refer to \gls{acr:per} as most practically relevant sampling strategy, we do not claim that DDQN with \gls{acr:per}, as proposed in the original \gls{acr:per} paper \citet{schaul_prioritized_2015}, represents the state-of-the-art in solving RL problems overall. Rather, we claim that to this day, no transition selection algorithm within experience replay has demonstrated efficiency improvements comparable to those of \gls{acr:per}, without incurring significant computational overhead. This claim is supported by the fact that most state-of-the-art \gls{acr:rl} algorithms use PER, while other prioritization strategy are scarcely used within state-of-the-art \gls{acr:rl} algorithms \citep{panahi2024investigatinginterplayprioritizedreplay}.

To support this notion, we have compiled Table~\ref{tab:rl_algorithms}, which lists leading algorithms on the Atari benchmark since the introduction of \gls{acr:per}, and indicates whether they use experience replay and  \gls{acr:per}.

\setlength\tabcolsep{5pt}
\begin{table}[ht]
\centering
\begin{tabular*}{\linewidth}{@{\extracolsep{\fill}} llcl}
\hline
\textbf{Algorithm (Year)} & \textbf{Authors [Year]} & \textbf{Uses ER} & \textbf{Uses \gls{acr:per}} \\
\hline
Rainbow & \citet{hessel_rainbow_2017}         & Yes & Yes \\
Ape-X DQN & \citet{horgan_distributed_2018}        & Yes & Yes \\
MuZero & \citet{schrittwieser_mastering_2019}           & Yes & Yes, for the Atari benchmark \\
R2D2 & \citet{kapturowski_recurrent_2019}             & Yes & Yes \\
Go-Explore & \citet{ecoffet_go-explore_2019}       & No  & No \\
NGU & \citet{badia_never_2020}              & Yes & Yes \\
Agent57 & \citet{badia_agent57_2020}         & Yes & Yes \\
EfficientZero & \citet{ye_mastering_2021}    & Yes & Yes \\
Bigger, Better, Faster & \citet{schwarzer_bigger_2023}              & Yes & Yes \\
Dreamer-v3 & \citet{hafner_mastering_2023}       & Yes & No, but \gls{acr:per} boosts performance\footnotemark[2]\\
SR-SPR & \citet{doro_sample-efficient_2023}           & Yes & Yes \\
EfficientZero-v2 & \citet{wang_efficientzero_2024} & Yes & Yes \\
\hline
\end{tabular*}
\caption{Overview of state-of-the-art reinforcement learning algorithms, highlighting whether they utilize Experience Replay (ER) and Prioritized Experience Replay (PER).}
\label{tab:rl_algorithms}
\end{table}

The findings within Table~\ref{tab:rl_algorithms} indicate that all state-of-the-art \gls{acr:rl} algorithms that rely on experience replay also rely on \gls{acr:per}. The sole exception of this is Dreamer-v3 \citep{hafner_mastering_2023}, which relies on uniform sampling for ease of implementation, but explicitly states \gls{acr:per} to boost performance. This provides evidence that \gls{acr:per} remains the de-facto standard prioritized sampling strategy, and therefore represents a key point of reference for our study.

\footnotetext[2]{While the classic Dreamer-v3 algorithm does not use \gls{acr:per} but uniform sampling, the authors explicitly report \gls{acr:per} to improve performance. To directly quote \citet{hafner_mastering_2023}: 
"While prioritized replay \citep{schaul_prioritized_2015} is used by
some of the expert algorithms we compare to and we found it to also improve the performance of
Dreamer, we opt for uniform replay in our experiments for ease of implementation."}

\subsection{Systematic review of proposed alternatives}

While we discussed alternative prioritized sampling strategies to provide a comprehensive overview of related work, these methods have seen limited adoption and are not integrated into state-of-the-art reinforcement learning algorithms. We lay out potential reasons for the sparse adoption of the approaches mentioned in the paper, and thereby discuss why we do not consider them a relevant baseline for the present paper.

While we have reviewed alternative prioritized sampling strategies to provide a comprehensive overview of related work, these methods have seen limited adoption and have not been widely integrated into state-of-the-art reinforcement learning algorithms. We outline possible factors contributing to their limited uptake and explain why, in the context of this study, we do not consider them appropriate baselines.

\emph{\cite{ramicic_entropy-based_2017}} proposed entropy-based sampling. Their evaluation focused on a single, non-standard environment and did not include a comparison against \gls{acr:per}. To the best of our knowledge, the authors also did not release code, which may limit the ease of direct application.

\emph{\citet{gao_prioritized_2021}} proposed reward-based sampling and evaluated their approach in two environments, \textsc{FetchReach-v1} and \textsc{Pendulum-v0}. While these experiments provide useful insights, no evaluation was presented on more complex domains such as Atari games, making direct comparison with the original \gls{acr:per} study less straightforward. From a theoretical perspective, an emphasis on rewards could potentially bias the algorithm toward greedier behavior, which might pose challenges in more complex settings. To the best of our knowledge, code was not made publicly available, which may limit immediate applicability.

\emph{\citet{brittain_prioritized_2019}} proposed a refinement to \gls{acr:per} by propagating priorities back through the sequence of transitions. Their evaluation compared the approach to a proportional variant of \gls{acr:per} with a different parameterization than the $\alpha = .5, \beta = .5$ setting recommended in the original paper, which may have influenced baseline performance. To the best of our knowledge, the work was released as a preprint in 2019 but has not appeared in a peer-reviewed venue, making it more difficult to fully gauge the impact of the proposed method.

\emph{\citet{zha_experience_2019}} introduced Experience Replay Optimization, a dynamic prioritization approach, and evaluated it on eight continuous control environments using DDPG. Their method was compared against \gls{acr:per} and demonstrated improved performance, though not within the original experimental setting of the \gls{acr:per} paper. Key hyperparameters such as $\alpha$ and $\beta$ were not reported, and, to the best of our knowledge, the authors did not release code, which may limit the reproducibility and practical applicability of their results.

\emph{\citet{oh_learning_2021}} introduced the Neural Experience Replay Sampler (NERS), which frames sample selection as a reinforcement learning problem by training a separate agent. While this approach is conceptually appealing, it introduces notable computational overhead, which may affect its practicality. The evaluation was conducted on Atari games for $100{,}000$ timesteps, rather than the conventional $50{,}000{,}000$, providing insights into the early stages of learning. The authors report improvements over \gls{acr:per} in this regime; however, details on the \gls{acr:per} configuration are not provided, and, to the best of our knowledge, the implementation has not been released, which may limit reproducibility.

\section{Detailed formal analysis}\label{ssec:detailed_formal_analysis}

We subsequently theoretically explore the properties of \gls{acr:reaper}. This section extends the formal analysis in Section~\ref{ssec:formal_analysis}.

\subsection{Convergence behavior}\label{sssec:convergence_behavior}

In the following, we provide a formal motivation for \gls{acr:reaper} by analyzing the influence of target bias on convergence behavior. We do so by showing that a misaligned target may degrade the value function, and then provide a decomposition of the expected error update.
 
\begin{lemma}[Update misalignment due to target bias]
\label{lemma:misalignment}
Let \( \mathbf{g}_t = \nabla_\theta (Q(S_t, A_t) - Q_{\text{target}}(S_t))^2 \) denote the gradient of the \gls{acr:tde} loss and let \( \mathbf{g}_t^\star = \nabla_\theta (Q(S_t, A_t) - Q^\star(S_t, A_t))^2 \) be the ideal gradient that aligns with the true value error. Then,
\begin{equation}
\langle \mathbf{g}_t, \mathbf{g}_t^\star \rangle = 2(Q(S_t, A_t) - Q^\star(S_t, A_t))^2 - 2(Q(S_t, A_t) - Q^\star(S_t, A_t))\varepsilon_t.
\end{equation}
\end{lemma}

\begin{proof}[Proof of Lemma~\ref{lemma:misalignment}]
We compute the gradients explicitly. We define
\begin{equation}
e_t := Q(S_t, A_t) - Q^\star(S_t, A_t), \quad \varepsilon_t := Q_{\text{target}}(S_t) - Q^\star(S_t, A_t).
\end{equation}
We now may rewrite the \gls{acr:tde},
\begin{equation}
\delta_t = Q_{\text{target}}(S_t) - Q(S_t, A_t) = (Q^\star(S_t, A_t) + \varepsilon_t) - Q(S_t, A_t) = -e_t + \varepsilon_t.
\end{equation}
Now, the gradients are
\begin{equation}
\mathbf{g}_t = 2(Q(S_t, A_t) - Q_{\text{target}}(S_t)) \nabla_\theta Q = 2(-\delta_t) \nabla_\theta Q(S_t, A_t),
\end{equation}
\begin{equation}
\mathbf{g}_t^\star = 2(Q(S_t, A_t) - Q^\star(S_t, A_t)) \nabla_\theta Q(S_t, A_t) = 2e_t \nabla_\theta Q(S_t, A_t).
\end{equation}
Hence,
\begin{equation}
\langle \mathbf{g}_t, \mathbf{g}_t^\star \rangle = 4(-\delta_t)e_t \|\nabla_\theta Q(S_t,A_t)\|^2 = 4(e_t - \varepsilon_t)e_t \|\nabla_\theta Q(S_t, A_t)\|^2.
\end{equation}
Simplifying yields
\begin{equation}
\langle \mathbf{g}_t, \mathbf{g}_t^\star \rangle = 4(e_t^2 - e_t\varepsilon_t) \|\nabla_\theta Q(S_t, A_t)\|^2,
\end{equation}
which proves the result up to a constant factor of the norm.
\end{proof}

This result shows that even when the \gls{acr:tde} is large, its usefulness critically depends on the reliability of the target value. When \( \varepsilon_t \) is large, the update may not improve the value function estimation. When \( \varepsilon_t \) is sign-misaligned with the current true estimation error $e_t$, the update will even degrade the value function, pushing \( Q(S_t,A_t) \) further away from \( Q^\star(S_t,A_t) \).

Based on these considerations, we proceed to compare various sampling strategies by analyzing the expected change in the squared Q-value error caused by a single update step. The following lemma provides a decomposition of this change and builds the foundation of our main theoretical result.

\begin{lemma}[Expected error update under sampling strategy \( \mu \)]
\label{lemma:update_expectation}
Let \( e_t = (Q(S_t, A_t) - Q^\star(S_t, A_t)) \). 
Let $Q$ denote the Q-function before an update, and let $Q'$ denote the Q-function after the update. Let $\mathbbm{E}_{\mu}[\Delta \|Q(S_t, A_t) - Q^\star(S_t, A_t)\|^2] = \mathbbm{E}_\mu[||Q'(S_t, A_t)-Q^\star(S_t, A_t)||^2 - ||Q(S_t, A_t)-Q^\star(S_t, A_t)||^2]$. Then, 
\begin{align}
\mathbb{E}_{\mu}\big[\Delta \|Q(S_t, A_t) - Q^\star(S_t, A_t)\|^2\big] 
&=2\eta \sum_{t=1}^{n} \mu_t \mathbb{E}\big[(Q(S_t, A_t) - Q^\star(S_t, A_t))\varepsilon_t\big]\nonumber\\
&\quad + \eta^2 \sum_{t=1}^{n} \mu_t \mathbb{E}[\delta_t^2] 
- 2\eta \sum_{t=1}^{n} \mu_t \mathbb{E}[e_t^2].
\end{align}
\end{lemma}

\begin{proof}[Proof of Lemma~\ref{lemma:update_expectation}]
We analyze the Q-value update
\begin{equation}
Q'(S_t, A_t) = Q(S_t, A_t) + \eta \delta_t.
\end{equation}
After the update
\begin{equation}
Q'(S_t, A_t) - Q^\star(S_t,A_t) = Q(S_t, A_t) + \eta \delta_t - Q^\star(S_t,A_t) = e_t + \eta \delta_t,
\end{equation}
the squared error becomes
\begin{equation}
(Q'(S_t, A_t) - Q^\star(S_t,A_t))^2 = (e_t + \eta \delta_t)^2 = e_t^2 + 2\eta e_t \delta_t + \eta^2 \delta_t^2.
\end{equation}
The expectation under sampling distribution \( \mu \) is
\begin{equation}
\mathbbm{E}[\Delta e_t] = \eta^2 \mathbbm{E}[\delta_t^2] + 2\eta \mathbbm{E}[e_t \delta_t].
\end{equation}
Note that \( \delta_t = Q_{\text{target}}(S_t) - Q(S_t,A_t) = \varepsilon_t - e_t \), so
\begin{equation}
e_t \delta_t = e_t (\varepsilon_t - e_t) = e_t \varepsilon_t - e_t^2,
\end{equation}
hence
\begin{equation}
\mathbbm{E}[\Delta e_t] = \underbracket[0.5pt]{\eta^2 \mathbbm{E}[\delta_t^2]}_{\text{(1)}} - \underbracket[0.5pt]{\mathbbm{E}[e_t^2])}_{\text{(2)}} + \underbracket[0.5pt]{2\eta (\mathbbm{E}[e_t \varepsilon_t]}_{\text{(3)}}.
\end{equation}
Summing over all transitions with \( \mu_t \) gives the result.
\end{proof}

This decomposition highlights three components: (1) the variance of the \gls{acr:tde}, (2) the true squared error and (3) the bias-error-interaction. The latter is key to explaining why ReaPER outperforms other sampling strategies.

A key factor in the reliability of bootstrapped targets is the extent of downstream \glspl{acr:tde}. Intuitively, if future states still exhibit significant \glspl{acr:tde}, the bootstrapped target for the current state is more likely to be biased. This motivates the following technical assumption.

\subsection{Discussion of Assumption \ref{assumption:bias}}\label{sssec:assumption}

Assumption~\ref{assumption:bias} establishes a relationship between the target bias for a given transition and the sum of \glspl{acr:tde} for downstream transitions. 
This aligns with a conventional perspective in TD-learning analysis, wherein bootstrapped targets predominantly inherit bias from inaccuracies in future value estimates.


Although Assumption~\ref{assumption:bias} appears rather limiting at first sight, it is in fact less strict than assumptions made in classical Q-learning analyses: classical Q-learning convergence proofs (see, e.g., Watkins \& Dayan, 1992) rely on global exploration assumptions, ensuring that every state-action pair is visited infinitely often, and on decaying learning rates to control noise.  
In contrast, Assumption~\ref{assumption:bias} takes a more local view, postulating that the target bias along an observed trajectory can be bounded by unresolved downstream \glspl{acr:tde}.  
While classical assumptions ensure eventual global accuracy, our assumption focuses on bounding the bias during finite-sample learning along actual agent trajectories, which is more aligned with practical deep \gls{acr:rl} settings.

Under Assumption~\ref{assumption:bias}, the reliability score $\mathcal{R}_t$ — which measures the proportion of downstream \gls{acr:tde} along a trajectory — provides an upper bound on the normalized target bias $\varepsilon_t$.  
Lemma~\ref{lemma:reliability} formalizes this relationship.

\subsection{Proof of Lemma~\ref{lemma:reliability}}
\begin{proof}\label{sssec:reliability_bias_bound}

From the definition of \( \mathcal{R}_t \), we have
\begin{equation}
1 - \mathcal{R}_t = \frac{\sum_{i = t+1}^{n} \delta_i^+}{\sum_{i = 1}^{n} \delta_i^+}.
\end{equation}
Multiplying both sides by \( \sum_{i = 1}^n \delta_i^+ \), we obtain
\begin{equation}
\sum_{i = t+1}^{n} \delta_i^+ = (1 - \mathcal{R}_t) \cdot \sum_{i = 1}^{n} \delta_i^+.
\end{equation}
Substituting this into Assumption~\ref{assumption:bias}, we find
\begin{equation}
|\varepsilon_t| \le \lambda \sum_{i = t+1}^{n} \delta_i^+ = \lambda (1 - \mathcal{R}_t) \cdot \sum_{i = 1}^{n} \delta_i^+,
\end{equation}
which proves the first inequality.

Rearranging the result gives
\begin{equation}
\mathcal{R}_t \le 1 - \frac{|\varepsilon_t|}{\lambda \sum_{i = 1}^{n} \delta_i^+},
\end{equation}
completing the proof.
\end{proof}

Moreover, it follows that
\begin{equation}
\mathcal{R}_t \le 1 - \frac{|\varepsilon_t|}{\lambda \sum_{i = 1}^n \delta_i^+}.
\end{equation}

Lemma~\ref{lemma:reliability} provides a formal link between the reliability score \( \mathcal{R}_t \) used in ReaPER and the target bias \( \varepsilon_t \). Under Assumption~\ref{assumption:bias}, transitions with large downstream \glspl{acr:tde} (i.e., large \( \sum_{i = t+1}^n \delta_i^+ \)) likely suffer from higher target bias. This justifies using \( \mathcal{R}_t \) to down-weight transitions with unreliable \glspl{acr:tde} in the sampling distribution as long as downstream transitions suffer from high \gls{acr:tde}. Consequently, ReaPER not only emphasizes transitions with high learning potential (large \( \delta_t^+ \)) but also prioritizes those with more reliable target estimates.

\subsection{Proof of Proposition~\ref{prop:hierarchy}}\label{sssec:hierarchy}
\begin{proof}
From Lemma~\ref{lemma:update_expectation}, the expected change in squared error per update, conditioned on the current Q-function, is
\begin{equation}
\Delta E = \eta^2 \sum_{t=1}^n \mu_t \mathbbm{E}[\delta_t^2] - 2\eta \sum_{t=1}^n \mu_t \mathbbm{E}[e_t^2] + 2\eta \sum_{t=1}^n \mu_t \mathbbm{E}[e_t \varepsilon_t].
\end{equation}
We seek to maximize the second term (true error reduction) while minimizing the third term (bias-error-interaction). The analysis proceeds by comparing the terms under the different sampling strategies. 



We now compare three sampling strategies:
    
\emph{Uniform sampling} (\( \mu_t = 1/n \)):  
No prioritization occurs. Transitions with small \( e_t^2 \) and potentially large \( e_t \varepsilon_t \) are sampled proportionally to their frequency of occurrence in the buffer. Hence, both the error reduction term \( \sum_t \mu_t \mathbbm{E}[e_t^2] \) and the bias term \( \sum_t \mu_t \mathbbm{E}[e_t \varepsilon_t] \) are solely determined by the buffer content, and sampling does not reduce error or bias.


\emph{PER sampling} (\( \mu_t \propto (\delta_t^+) \)):  
PER prioritizes transitions with large \gls{acr:tde} \( \delta_t^+ \), which correlates with larger \( e_t^2 \). As such, the sampling increases error reduction over buffer content, 
\begin{equation}
\sum_{t=1}^n \mu_t^{\mathrm{PER}} \mathbbm{E}[e_t^2] \gg \sum_{t=1}^n \mu_t^{\mathrm{Uniform}} \mathbbm{E}[e_t^2],
\end{equation}
leading to faster true error reduction compared to uniform sampling.  
However, PER does not account for target bias \( \varepsilon_t \). Nevertheless, since PER focuses updates on transitions with large \glspl{acr:tde} (rather than arbitrary ones), it slightly reduces the bias-error-interaction compared to uniform:
\begin{equation}
\sum_{t=1}^n \mu_t^{\mathrm{PER}} \mathbbm{E}[e_t \varepsilon_t] \ll \sum_{t=1}^n \mu_t^{\mathrm{Uniform}} \mathbbm{E}[e_t \varepsilon_t].
\end{equation}

\emph{ReaPER sampling} (\( \mu_t \propto \mathcal{R}_t \delta_t^+ \)):  
Prioritizes transitions with large \gls{acr:tde} \emph{and} high reliability \( \mathcal{R}_t \), thus additionally considering target bias (see Lemma~\ref{sssec:reliability_bias_bound}).
Thus, ReaPER achieves
\begin{equation}
\sum_{t=1}^n \mu_t^{\mathrm{ReaPER}} \mathbbm{E}[e_t^2] \gg \sum_{t=1}^n \mu_t^{\mathrm{PER}} \mathbbm{E}[e_t^2] \gg \sum_{t=1}^n \mu_t^{\mathrm{Uniform}} \mathbbm{E}[e_t^2],
\end{equation}
for the true error term, and
\begin{equation}
\sum_{t=1}^n \mu_t^{\mathrm{ReaPER}} \mathbbm{E}[e_t \varepsilon_t] \ll \sum_{t=1}^n \mu_t^{\mathrm{PER}} \mathbbm{E}[e_t \varepsilon_t] \ll \sum_{t=1}^n \mu_t^{\mathrm{Uniform}} \mathbbm{E}[e_t \varepsilon_t],
\end{equation}
for the bias term.

Therefore, ReaPER leads to the steepest expected decrease in squared error per update, followed by PER, followed by uniform sampling. Summing the per-step improvements over training steps, we conclude
\begin{equation}
\mathbbm{E}_{\mu^{\mathrm{Uniform}}}[\|Q_T - Q^\star\|^2] \geq \mathbbm{E}_{\mu^{\mathrm{PER}}}[\|Q_T - Q^\star\|^2] \geq \mathbbm{E}_{\mu^{\mathrm{ReaPER}}}[\|Q_T - Q^\star\|^2],
\end{equation}
as claimed.
\end{proof}

\subsection{Discussion of Remark~\ref{remark:suboptimal}}\label{sssec:remark} While Assumption~\ref{assumption:bias} is stated under the premise that the agent follows an optimal policy, it can be extended to fixed but suboptimal policies. We now briefly outline the modifications necessary if the agent follows a fixed but suboptimal policy. In this case, the target bias \( \varepsilon_t \) cannot be bounded solely by unresolved downstream \glspl{acr:tde}, as suboptimal actions introduce an additional, trajectory-independent bias. Formally, suppose there exists a constant \( \zeta \geq 0 \) such that for all transitions along observed trajectories,
\begin{equation}
|Q^\pi(S_t, A_t) - Q^\star(S_t, A_t)| \leq \zeta,
\end{equation}
where \( Q^\pi \) denotes the action-value function under the fixed policy \( \pi \). Then, under analogous reasoning to Assumption~\ref{assumption:bias}, the target bias satisfies
\begin{equation}
|\varepsilon_t| \leq \lambda \sum_{i = t+1}^n \delta_i^+ + \zeta.
\end{equation}
This adjusted bound propagates through the subsequent results. In particular, Lemma~\ref{sssec:reliability_bias_bound} becomes
\begin{equation}
|\varepsilon_t| \leq \lambda (1 - \mathcal{R}_t) \sum_{i=1}^n \delta_i^+ + \zeta,
\end{equation}
and the reliability bound adjusts accordingly. In the error decomposition of Lemma~\ref{lemma:update_expectation} and the convergence hierarchy in Proposition~\ref{sssec:hierarchy}, the additive term \( \zeta \) introduces a bias floor that does not vanish through learning. Consequently, while ReaPER still achieves improved sampling efficiency by reducing the impact of target misalignment, the achievable Q-function accuracy is ultimately lower-bounded by \( \zeta \). In the limit as \( \zeta \to 0 \) (the policy approaches optimality), we recover the original theory.

\subsection{Proof of Proposition~\ref{prop:variance:reduction}}\label{sssec:variance_reduction}

\begin{proof}  
First, we note that we can assume without loss of generality that there is a $\tau>0$ such that the distribution $\mu^\star$ satisfies 
\begin{equation} \label{eq:wlog:pf:sampling:dist}
    \sum_{i=1}^N \mu^\star_i \delta_i^+ \geq \tau.
\end{equation}
If such a $\tau>0$ does not exist, that implies $\sum_{i=1}^N \mu^\star_i \delta_i^+$ which only holds if $\delta_i = 0$ for all $i=1,\dots,N$ -- a setting in which any sampling distribution from the buffer is optimal. Hence, equipped with \eqref{eq:wlog:pf:sampling:dist}, the optimal sampling distribution $\mu^\star$ is characterized as a solution to the following optimization problem
\begin{align}
\min_{\mu \in \Delta_N} & \quad \sum_{i=1}^N \mu_i \sigma_i^2 \\
\text{subject to} & \quad \sum_{i=1}^N \mu_i \delta_i^+ \ge \tau.
\end{align}
We introduce Lagrange multipliers \( \lambda \ge 0 \) for the inequality constraint and \( \nu \) for the probability normalization constraint. Then, the Lagrangian reads
\begin{equation}
\mathcal{L}(\mu, \lambda, \nu) = \sum_{i=1}^N \mu_i \sigma_i^2 + \lambda \left( \tau - \sum_{i=1}^N \mu_i \delta_i^+ \right) + \nu \left( 1 - \sum_{i=1}^N \mu_i \right)
\end{equation}
The KKT conditions for optimality are:
\begin{align}
\text{(Stationarity)} &\quad \frac{\partial \mathcal{L}}{\partial \mu_i} = \sigma_i^2 - \lambda \delta_i^+ - \nu = 0 \quad \text{for all } i=1,\dots,N \\
\text{(Primal feasibility)} &\quad \sum_{i=1}^N \mu_i \delta_i^+ \ge \tau, \quad \sum_{i=1}^N \mu_i = 1, \quad \mu_i \ge 0 \\
\text{(Dual feasibility)} &\quad \lambda \ge 0 \\
\text{(Complementary slackness)} &\quad \lambda \left( \tau - \sum_{i=1}^N \mu_i \delta_i^+ \right) = 0
\end{align}
In the following we distinguish two cases.

\paragraph{Case 1:}
Suppose \( \sum_{i=1}^N \mu_i \delta_i^+ > \tau \). Then, complementary slackness implies \( \lambda = 0 \). The stationarity condition becomes
\begin{equation}
\sigma_i^2 - \nu = 0 \quad \Rightarrow \quad \sigma_i^2 = \nu \quad \text{for all } i=1,\dots,N,
\end{equation}
which is only possible if all \( \sigma_i^2 \) are equal. In general, this is not the case, so the constraint must be active at optimality.

\paragraph{Case 2:} 
Then \( \sum_{i=1}^N \mu_i \delta_i^+ = \tau \), and complementary slackness implies \( \lambda > 0 \). From the stationarity condition, we obtain
\begin{equation}
\sigma_i^2 - \lambda \delta_i^+ - \nu = 0 \quad \Rightarrow \quad \sigma_i^2 = \lambda \delta_i^+ + \nu
\end{equation}
Solving for \( \lambda \), we get
\begin{equation}
\lambda = \frac{\sigma_i^2 - \nu}{\delta_i^+}
\end{equation}
This must hold for all \( i=1,\dots,N \), so the right-hand side must be constant across \( i \), which implies
\begin{equation}
\frac{\sigma_i^2}{\delta_i^+} - \frac{\nu}{\delta_i^+} = \text{constant}
\quad \overset{\text{(a)}}{\Rightarrow} \quad \mu_i^\star \propto \frac{\delta_i^+}{\sigma_i^2}
\end{equation}
To justify the implication (a), note that the stationarity condition directly states
$\sigma_i^2 - \nu = \lambda \delta_i^+ \Rightarrow \frac{1}{\lambda} = \frac{\delta_i^+ }{\sigma_i^2 - \nu}$. Therefore, a higher value $\frac{\delta_i^+ }{\sigma_i^2 - \nu}$ implies a higher gradient contribution exactly where $\mu_i$ should be large.
 \end{proof}

\section{Empirical validation of Assumption~\ref{assumption:bias}}\label{sec:assumption_testing}

To empirically support Assumption~\ref{assumption:bias},  we conducted a simulation leveraging a \textsc{BlindCliffwalk} experiment. The \textsc{BlindCliffwalk} environment is a stylized \gls{acr:rl} setting introduced in the original \gls{acr:per} paper \citep{schaul_prioritized_2015}, which we adopt to enable comparability with prior work. In this environment, an agent has to make $n$ consecutive correct binary decisions to obtain a reward of $1$. If the incorrect action in a given state is selected, the agent is reset to the initial state, incurring a reward of 0. In this controlled environment, the true Q-values can be computed exactly, enabling direct empirical evaluation of our theoretical bound. We evaluated both a \gls{acr:tql} setup and a simple \gls{acr:dqn} with a single 16-neuron hidden layer. Each algorithm was evaluated along four different cliffwalk lengths $(4, 6, 8, 10)$ for $100$ iterations, with $5{,}000$ episodes of training time.

We report average rewards as an indicator of convergence. We further report observed bound violations, defined as 

\begin{equation}
\lvert\varepsilon_t\rvert > \sum_{i = t+1}^{n} \delta_i^+.
\end{equation}

Additionally, we report the empirically observed $\lambda$, defined as

\begin{equation}
\lambda = \frac{\lvert\varepsilon_t\rvert}{\sum_{i = t+1}^{n} \delta_i^+}.
\end{equation}

\begin{figure}[t!]
    \centering
    \includegraphics[width=1\linewidth]{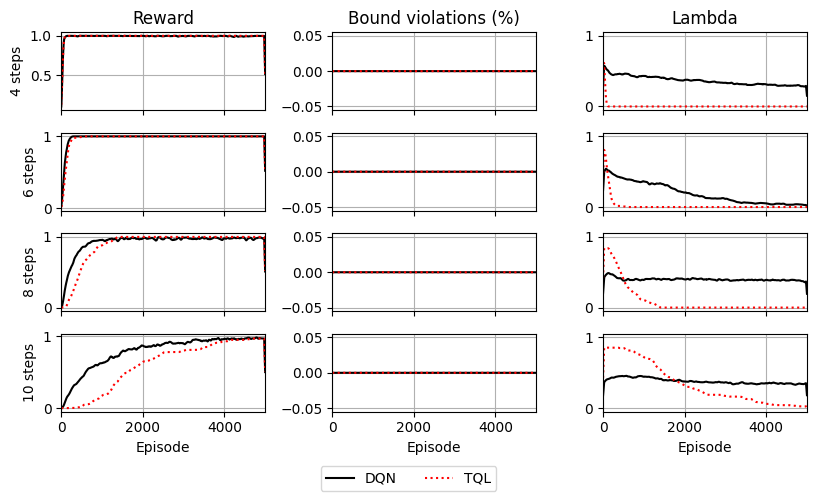}
    \caption{Rewards, bound violations and $\lambda$ in a 4-, 6-, 8-, and 10-step \textsc{BlindCliffwalk} environment for \gls{acr:tql} and \gls{acr:dqn}, averaged across 100 runs and smoothened across 50 consecutive data points.}
    \label{fig:assumption-testing}
\end{figure}

As shown in Figure~\ref{fig:assumption-testing}, our results indicate that the bound established in Assumption~\ref{assumption:bias} is never violated, neither in the \gls{acr:tql} nor in the \gls{acr:dqn} setting. Moreover, we observe that $\lambda$ decreases as the model converges. This trend is especially pronounced in the tabular setting, yet also apparent in the \gls{acr:dqn} setting, and provides empirical support for our theoretical claim that the bound tightens as the agent's policy improves. Taken together, these findings suggest that the bound is reliable, even under complex non-tabular function-approximation dynamics.

\setlength\tabcolsep{5pt}
\begin{table}
\centering
\begin{tabular*}{\linewidth}{@{\extracolsep{\fill}} lcc}
\hline
Parameter & \textsc{TQL} & \textsc{DQN} \\
\hline
Learning rate & 0.5 & 0.01 \\
Gamma & 0.99 & 0.99 \\
Exploration &  Random tie break & $\epsilon=1$ until first reward, then $\epsilon=0$ \\
Buffer size & - & 1000 \\
Batch size & 1 & 10 \\
\hline
\end{tabular*}
\caption{Hyperparameters for the \gls{acr:tql} and \gls{acr:dqn} conditions for empirical testing of Assumption 3.4 in the \textsc{BlindCliffwalk} environment.}
\label{tab:tql_dqn_hyperparams}
\end{table}

\section{Convergence in a stylized setting}\label{ssec:stylized_setting}

We consider a single episode within a stylized setting. The agent is following the optimal path, $Q_{\text{target}}(S_{\idxPrimary}) = Q(S_{\idxPrimary+1}, A_{\idxPrimary+1})$ where $A_{\idxPrimary+1}= \pi^\star(S_{\idxPrimary+1})$ for all $\idxPrimary \in \{1, \dots, n-1\}$. At the end of this episode, the agent obtains a final reward $R_n = 1$. There are no intermediary rewards. The agent aims to learn the correct Q-values for all transitions within this trial using a \gls{acr:tql} approach \citep{watkins_q-learning_1992}. $Q_{\text{target}}(S_{\idxPrimary})$ are continuously updated to be $Q(S_{\idxPrimary+1},A_{\idxPrimary+1})$, with $Q_{\text{target}}(S_n)$ being 1. We consider a transition $C_{\idxPrimary}$ to be learned if the Q-value reaches its (for real applications mostly unknown) ground-truth Q-value, $Q(S_{\idxPrimary},A_{\idxPrimary}) = Q^\star(S_{\idxPrimary},A_{\idxPrimary})$. We consider the model to have converged when all Q-values reach their ground-truth Q-value, $Q(S_{\idxPrimary},A_{\idxPrimary}) = Q^\star(S_{\idxPrimary},A_{\idxPrimary})$ for $\idxPrimary \in \{1,\dots,n\}$. This is the case when all \glspl{acr:tde} within this trial have a value of 0, i.e., $\delta_{\idxPrimary} = 0$ for $\idxPrimary \in \{1,\dots,n\}$. 

The agent learns by repeatedly selecting $k$ transition indices. For every selected transition $C_{\idxSelected}$, the Bellman equation is solved to update the Q-value, $Q(S_{\idxSelected}, A_{\idxSelected}) \gets Q(S_{\idxSelected}, A_{\idxSelected}) + \gamma \cdot \delta_{\idxSelected}$. For the sake of simplicity, we assume a discount factor $\gamma = 1$, a batch size $k = 1$, and a learning rate $\eta = 1$. As $\eta = 1$, learning on a transition $C_\idxSelected$ implies setting the Q-value to $Q_{\text{target}}$, i.e., $Q(S_{\idxSelected}, A_{\idxSelected}) = Q_{\text{target}}(S_{\idxSelected})$. We repeat this iterative process of sampling and the respective Q-value adaptation until the model converged.

We use three different selection strategies to identify the transition index $\idxSelected$, which determines the transition $C_{\idxSelected}$ that the model trains on next, uniform sampling, \gls{acr:perg}, and \gls{acr:reaperg}. Uniform sampling selects transitions at random with equal probability. \gls{acr:perg} selects the transition with the highest \gls{acr:tde} $\delta$. \gls{acr:reaperg} selects the transition with the highest reliability-adjusted \gls{acr:tde} $\Psi$. in both \gls{acr:perg} and \gls{acr:reaperg}, if there is no unique maximum, ties are resolved by random choice.

We compare these selection strategies to the optimal solution, the \emph{Oracle}. The Oracle selects the transition $C_\idxOracle$ with the highest index that has a absolute \gls{acr:tde} greater than zero, $\idxOracle= \max(\idxPrimary \mid \delta_{\idxPrimary} \neq 0)$ for $\idxPrimary \in \{1, \dots, n\}$. Given $\eta = 1$, using the Oracle, the agent will always converge within $\sum_{\idxPrimary = 1}^{n}\mathbbm{1}_{\delta_{\idxPrimary} \neq 0} \leq n$ steps and is therefore optimal. We compare the sampling strategies under varying levels of $Q_{\text{target}}$ reliability. $Q_{\text{target}}$ reliability here is determined by the extent of target Q-values for unlearned transitions without immediate reward varying from zero: Reliability is high if $Q_{\text{target}}$ values remains close to the immediate observed reward unless specifically learned otherwise. $Q_{\text{target}}$ reliability decreases with more $Q_{\text{target}}$ values deviating from zero without observation of an immediate reward and without being explicitly learned. In reality, this may happen either through Q-value initialization or - more importantly, when using Q-functions - erroneous Q-value generalization across state-action pairs. 

In the present example, we simulate different levels of $Q_{\text{target}}$ reliability using different $Q$-value initializations. Specifically, we consider three $Q_{\text{target}}$ reliability conditions: High, medium and low $Q_{\text{target}}$ reliability. In all conditions, all $Q(S_{\idxPrimary}, A_{\idxPrimary})$ for $\idxPrimary \in \{1,\dots,n\}$ are first initialized to zero. Then, depending on the reliability condition, some of these initializations are overwritten with ones to induce unreliability. In the \textit{low reliability} condition, every second Q-value is overwritten. In the \textit{medium reliability} condition, every fourth Q-value is overwritten. In the \textit{high reliability} condition, no value is overwritten. 

We employ every selection mechanism to train until convergence for episodes of length 10 to 100 across all reliability conditions. As shown in Figure 2, uniform sampling converges the slowest across all reliability conditions. \gls{acr:perg} finds the optimal transition selection order when target reliability is high. However, \gls{acr:perg}'s convergence speed quickly diminishes as $Q_{value}$ reliability decreases. \gls{acr:reaperg} on the other hand actively accounts for changes in $Q_{value}$ reliability. By doing so, it consistently finds the ideal solution regardless of $Q_{value}$ reliability.

\begin{figure}[t!]
    \centering
    \includegraphics[width=1\linewidth]{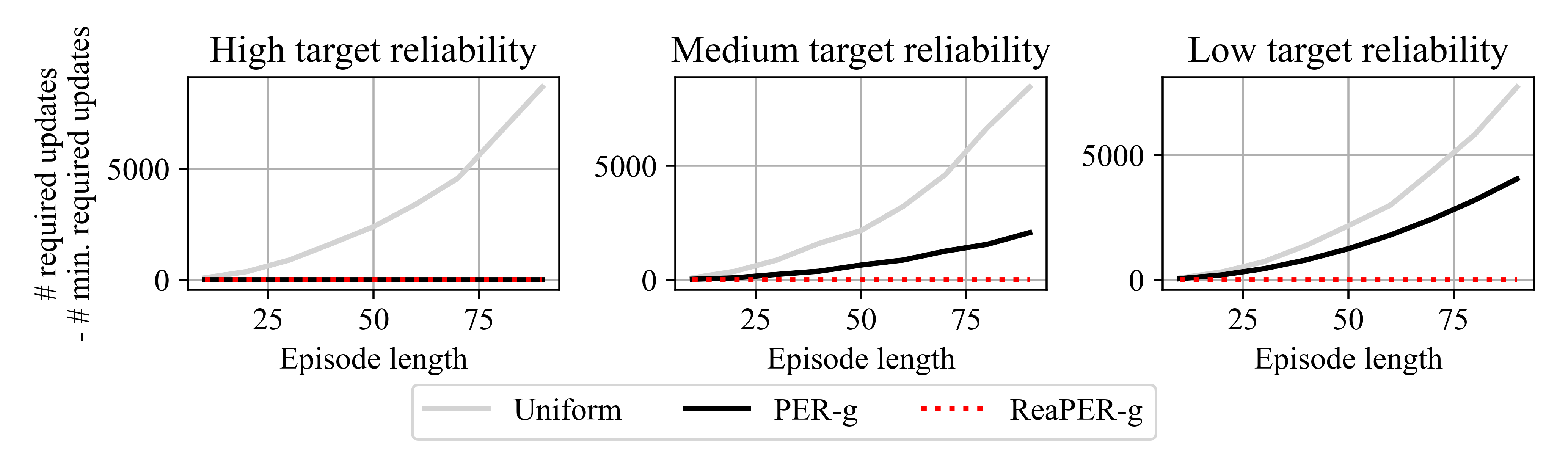}
    \caption{Performance comparison in a stylized setting for three sampling methods; uniform sampling, \gls{acr:perg}, and \gls{acr:reaperg}. Performance is quantified as the number of updates until convergence minus the minimally required number of updates given by an Oracle. Each sampling method is evaluated using different episode lengths (between 10 and 100) and different levels of $Q_{\text{target}}$ reliability (high, medium and low).}
    \label{fig:motivating-example}
\end{figure}

\section{Experimental settings}\label{ssec:experimental_settings}

All training parameters for all environments were set to according to pre-existing recommendations from previous research \citep{mnih_human-level_2015, schaul_prioritized_2015, van_hasselt_deep_2015, rl-zoo3}. The full experimental settings are subsequently described in detail to enable full reproducibility.

\paragraph{Experience replay parameters} Following the suggestion for proportional \gls{acr:per} for \gls{acr:ddqn} in \citet{schaul_prioritized_2015}, $\alpha$ was set to $0.6$ for \gls{acr:per}. As we expect increased need for regularization due to the reliability-driven priority scale-down, we expected values smaller values for $\alpha$ and $\omega$ in \gls{acr:reaper}. We performed minimal hyperparameter tuning to find a suitable configuration. We did so by training a single game, \textsc{Qbert}, on three configurations for $\alpha$ and $\omega$, (1) $\alpha=0.2$, $\omega=0.4$, (2) $\alpha=0.3$, $\omega=0.3$ and (3) $\alpha=0.4$, $\omega=0.2$. For the runs presented in the paper, the best-performing variant ($\alpha=0.4$, $\omega=0.2$) was used. For both \gls{acr:per} and \gls{acr:reaper}, $\beta$ linearly increased with training time, $\beta \gets (0.4 \to 1.0)$ as proposed in \citet{schaul_prioritized_2015}. A buffer size of $10^6$ was used. 

\paragraph{Atari preprocessing} As in \citet{mnih_human-level_2015}, Atari frames were slightly modified before being processed by the network. Preprocessing was performed using StableBaselines3's \textit{AtariWrapper} \citep{antonin_raffin_stable-baselines3_2024}. All image inputs were rescaled to an 84x84 grayscale image. After resetting the environment, episodes were started with a randomized number of \textit{NoOp}-frames (up to 30) without any operation by the agent, effectively randomizing the initial state and consequently preventing the agent from learning a single optimal path through the game. Four consecutive frames were stacked to a single observation to provide insight into the direction of movement. Additionally, a termination signal is sent when a life is lost. All these preprocessing steps can be considered standard practice for \textsc{Atari} games (e.g., \citet{mnih_playing_2013, mnih_human-level_2015, schaul_prioritized_2015}).

\paragraph{Training specifications} Hyperparameters for learning to play \textsc{CartPole}, \textsc{Acrobot} and \textsc{LunarLander} were set to RL Baselines3 Zoo recommendations \citep{rl-zoo3}. Hyperparameters for learning to play \textit{Atari} games are set based on previous research by \citet{schaul_prioritized_2015}, \citet{mnih_human-level_2015} and \citet{van_hasselt_deep_2015}.

\setlength\tabcolsep{5pt}
\begin{table}
\centering
\begin{tabular*}{\linewidth}{@{\extracolsep{\fill}} lcccc}
\hline
Parameter & \textsc{CartPole} & \textsc{Acrobot} & \textsc{LunarLander} & \textsc{Atari} \\
\hline
Learning rate & 2.3e-3 & 6.3e-4 & 6.3e-4 & 625e-5\footnotemark[3] \\

Budget in timesteps & 5e4 & 1e5 & 1e5 & 5e7 (2e8 frames)\\
Buffer size & 1e5 & 5e4 & 5e4 & 1e6 \\
Timestep to start learning & 1e3 & 1e3 & 1e3 & 5e4 \\
Target network update interval & 10 & 250 & 250 & 3e4\\
Batch size & 64 & 128 & 128 & 32 \\
Steps per model update & 256 & 4 & 4 & 4 \\
Number of gradient steps & 128 & 4 & 4 & 1 \\
Exploration fraction & 0.16 & 0.12 & 0.12 & 0.02 \\
Initial exploration rate & 1 & 1 & 1 & 1 \\
Final exploration rate & 0.04 & 0.1 & 0.1 & 0.01\\
Evaluation exploration fraction & 0.001 & 0.001 & 0.001 & 0.001\\
Number of evaluations & 100 & 100 & 100 & 200 \\
Trajectories per evaluation & 5 & 5 & 5 & 1 \\
Gamma & 0.99 & 0.99 & 0.99 & 0.99 \\
Max. gradient norm & 10 & 10 & 10 & $\infty$ \\
Reward threshold & 475 & -100 & 200 & $\infty$ \\
Optimizer & Adam & Adam & Adam & RMSprop \\
\hline
\end{tabular*}
\caption{Comprehensive documentation of hyperparameters used within the \textsc{CartPole}, \textsc{Acrobot}, \textsc{LunarLander} and \textsc{Atari} environments.}
\label{tab:hyperparams}
\end{table}

\footnotetext[3]{In the uniform experience replay condition, the learning rate was increased to 2.5e-4, as recommended in seminal work \citep{mnih_human-level_2015, schaul_prioritized_2015, van_hasselt_deep_2015}.}

\paragraph{Network architecture} For \textsc{CartPole}, \textsc{Acrobot} and \textsc{LunarLander}, the network architecture was equivalent to \textit{StableBaselines3}' default architecture \citep{antonin_raffin_stable-baselines3_2024}: A two-layered fully-connected net with 64 nodes per layer was used. For image observations as in the \textsc{Atari-10} <benchmark, the input was preprocessed to size $84 \times 84 \times 4$. It was then passed through three convolutional layers and two subsequent fully connected layers. The network architecture was equal to the network architecture used in \citet{mnih_human-level_2015}. Rectified Linear Units \citep{agarap_deep_2018} were used as the activation function.

\paragraph{Evaluation} For the environments \textsc{CartPole}, \textsc{Acrobot} and \textsc{LunarLander}, $100$ evaluations were evenly spaced throughout the training procedure. Each agent evaluation consisted of five full trajectories in the environment, going from initial to terminal state. The evaluation score of a single agent evaluation is the average total score across those five evaluation trajectories. Training was stopped when the agent reached a predefined reward threshold defined in the Gymnasium package \citep{towers_gymnasium_2024}. For \textsc{Atari} environments, as in \citet{schaul_prioritized_2015}, $200$ evaluations consisting of a single trajectory were evenly spaced throughout the training procedure. No reward threshold was set.

\textbf{Score normalization}: For \textsc{Atari} games, scores were normalized to allow for comparability between games. Let $\Xi_{raw}$ denote a single evaluation score that is to be normalized. Let $\Xi_{random}$ denote the score achieved by a randomly initialized policy in this game. Let $\Xi_{max}$ denote the highest score achieved in this game across either condition, \gls{acr:reaper} or \gls{acr:per}. The normalized score $\Xi_{norm}$ is then calculated via

\begin{equation}
\Xi_{norm} = \frac{\Xi_{raw} - \Xi_{random}}{\Xi_{high} - \Xi_{random}}.
\end{equation}

\textbf{Percentage improvement}:

For the environments \textsc{CartPole}, \textsc{Acrobot} and \textsc{LunarLander}, we compute percentage improvements in median timesteps-to-convergence $\mathcal{T}$ until meeting a specified reward threshold of \gls{acr:reaper} upon another condition $\mathcal{C}$, as stated within Section~\ref{sec:experiments},  as

\begin{equation}
\text{Improvement over } \mathcal{C} = \frac{\mathcal{T}_{\mathcal{C}} - \mathcal{T}_{\text{ReaPER}}}{\mathcal{T}_{\mathcal{C}}} \times 100.
\end{equation}

For the \textsc{Atari} environments, we compute percentage improvements in peak score $\mathcal{P}$ of \gls{acr:reaper} upon another condition $\mathcal{C}$ , as stated within Section~\ref{sec:experiments}, as

\begin{equation}
\text{Per-game improvement over } \mathcal{C} = \frac{\mathcal{P}_{\text{ReaPER}} - \mathcal{P}_{\mathcal{C}}}{\mathcal{P}_{\mathcal{C}}}\times 100.
\end{equation}

The overall improvement across the \textsc{Atari-10} benchmark is computed as the median of the improvements observed on each individual game.

\section{Atari-10 results}\label{ssec:atari10_results}

\begin{figure}[H]
    \centering
    \includegraphics[width=.75\linewidth]{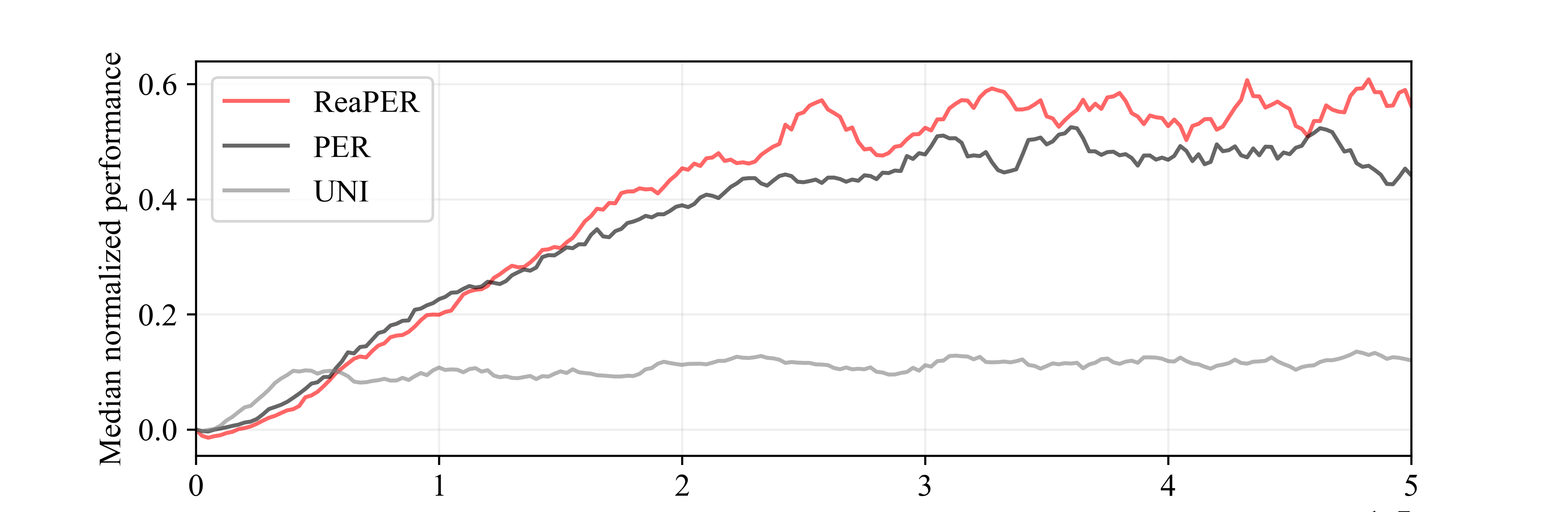}
    \caption{Median of normalized scores across the \textsc{Atari-10} benchmark for \gls{acr:reaper}, \gls{acr:per} and uniform experience replay (UNI) with a moving average smoothed over 10 points. }
    \label{fig:agg_performance_across_atari_noCUMMAX}
\end{figure}

Figure~\ref{fig:agg_performance_across_atari_noCUMMAX} displays the median of normalized scores across games gathered throughout 200 evaluation periods, which were evenly spaced-out throughout the 50 million training timesteps. 

\begin{figure}[H]
    \centering
    \includegraphics[width=1\linewidth]{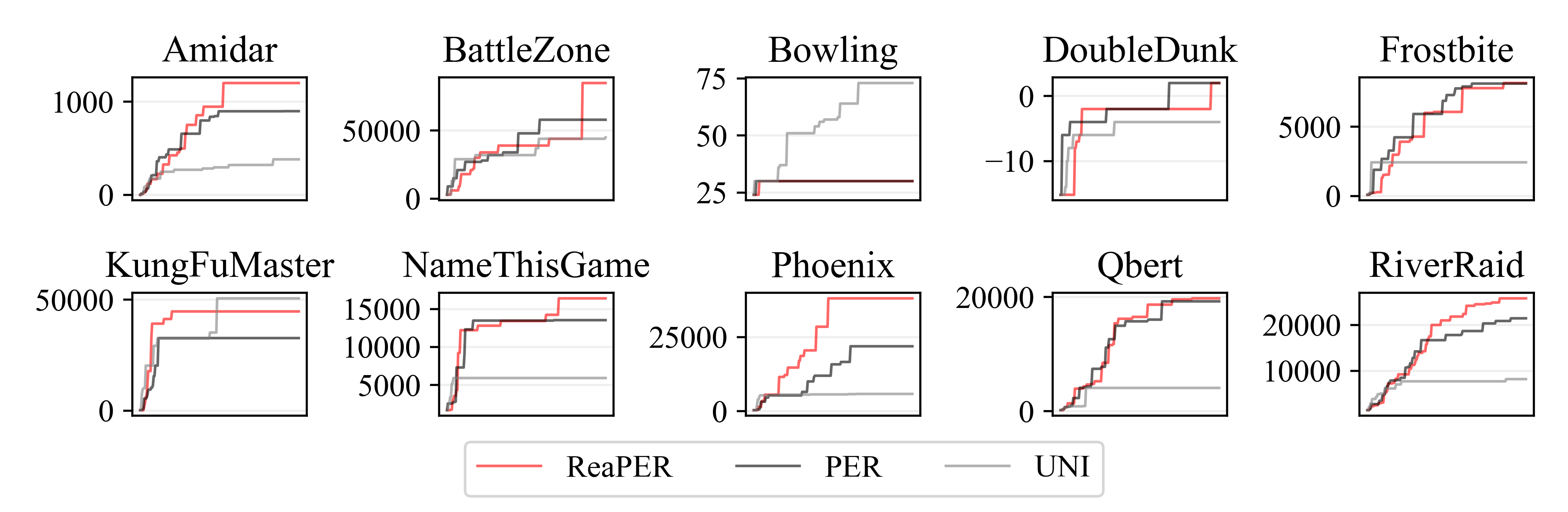}
    \caption{Cumulative maximum evaluation score per game from the \textsc{Atari-10} benchmark across the training period for \gls{acr:reaper}, \gls{acr:per} and uniform experience replay (UNI).}
    \label{fig:allatari_cummax}
\end{figure}

Figure~\ref{fig:allatari_cummax} displays the cumulative maximum scores per game gathered across 200 evaluation periods. 

\section{Partial observability}\label{ssec:partial_observability}

In Atari environments, a single-frame observation does not convey object velocities or movement directions, making the vanilla problem setting a partially observable \gls{acr:mdp} \citep{hausknecht2017deeprecurrentqlearningpartially}. Standard practice mitigates this by stacking four consecutive frames as the agent’s observation to ease the learning process \citep{mnih_human-level_2015}. To perform a deliberate study of performance under partial observability, we intentionally restrict the agent to a single-frame observation.

\begin{figure}[H]
    \centering
    \includegraphics[width=1\linewidth]{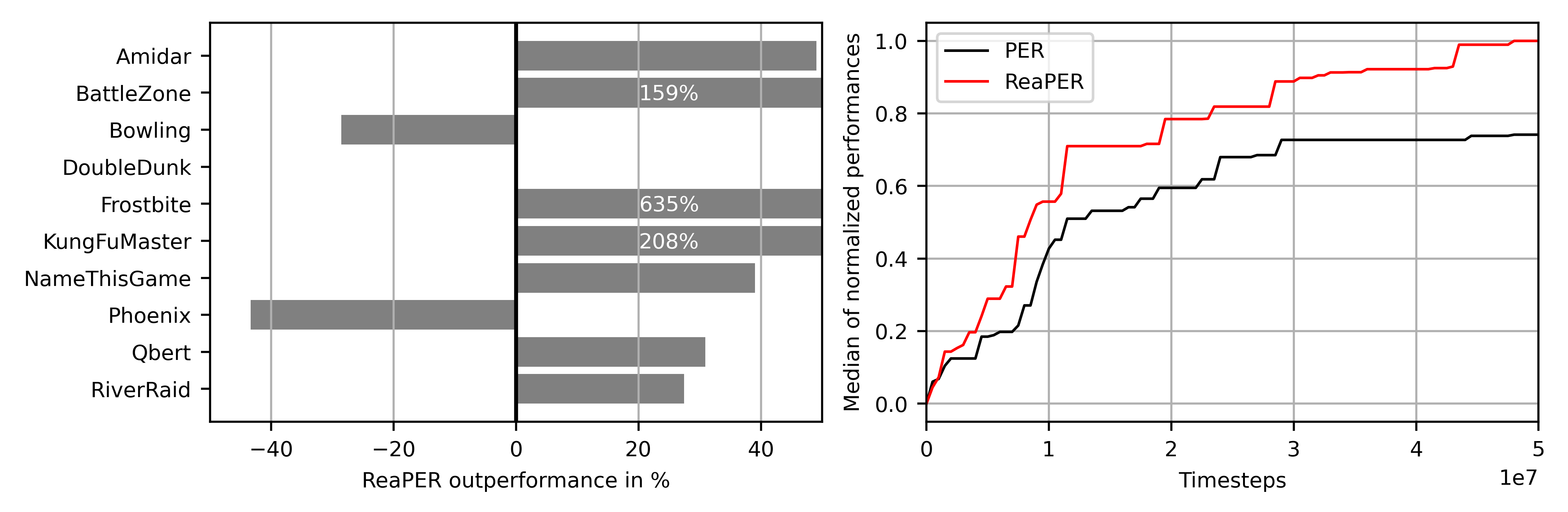}
     \caption{Left: Peak score increase of \gls{acr:reaper} over \gls{acr:per} under partial observability, induced by single-frame observations. Right: Median of the normalized cumulative maximum of scores across the Atari-10 benchmark for \gls{acr:reaper} and \gls{acr:per} under partial observability. The normalized score at timestep $\idxPrimary$ is calculated by dividing the difference between the current score and the random score by the difference between the maximum score in this game across all sampling strategies and the random score.}
    \label{fig:ablation_partial_observability}
\end{figure}

\begin{figure}[H]
    \centering
    \includegraphics[width=1\linewidth]{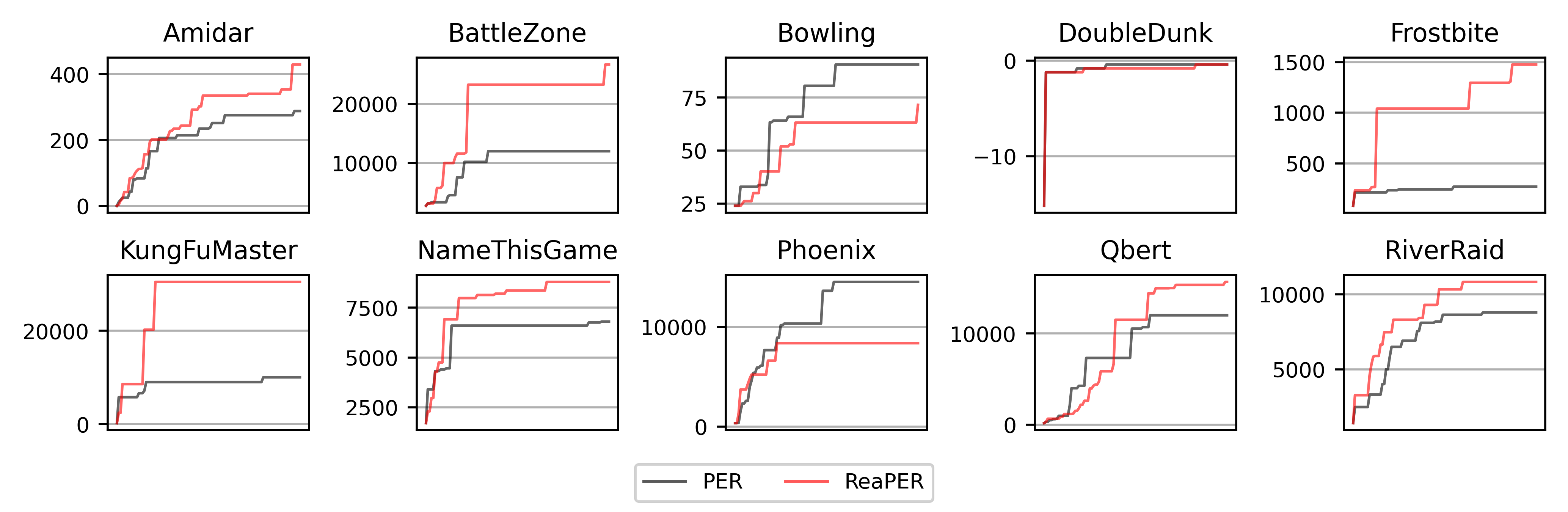}
    \caption{Cumulative maximum evaluation score per game from the \textsc{Atari-10} benchmark across the training period for \gls{acr:reaper} and \gls{acr:per} under partial observability, induced by single-frame observations.}
    \label{fig:allatari_cummax_partial_obs}
\end{figure}

Under this constraint, \gls{acr:reaper} outperforms \gls{acr:per} in 8 of 10 games, achieving a median relative improvement of $34.97\%$, substantially exceeding the $22.97\%$ gain observed in the fully observable setting. This result indicates that the benefits of reliability adjustment are further amplified under partial observability. Aggregated results are shown in Figure~\ref{fig:ablation_partial_observability}. Per-game curves are shown in Figure~\ref{fig:allatari_cummax_partial_obs}.

\section{Hardware specification}
The \textsc{Atari} experiments were conducted on a workstation equipped with an AMD Ryzen 9 7950X CPU (32 cores at 4.5 GHz), 128 GB of RAM, and an NVIDIA RTX 4090 GPU with 24 GB of memory (driver version 12.3). All other numerical experiments were performed on a 2024 MacBook Air with an Apple M3 processor.

\end{document}